%% file: main.tex
\documentclass{article} 
\usepackage{iclr2022_conference,times}

\input{math_commands.tex}

\usepackage{hyperref}
\usepackage{url}
\usepackage{algorithmic}

\usepackage{booktabs}
\usepackage{makecell}

\usepackage{lipsum}
\usepackage{comment}
\usepackage{multirow}
\usepackage{wrapfig}
\usepackage{algorithm}
\usepackage{algorithmic}
\usepackage{subfigure}

\usepackage{graphicx}
\usepackage{color}

\usepackage[english]{babel}
\usepackage{amsthm}
\definecolor{darkgreen}{rgb}{0,0.7,0}

\newcommand{\green}[1]{{\color{darkgreen}{#1}}}

\renewcommand{\L}{\mathcal{L}}

\renewcommand{\O}{\mathcal{O}}

\newcommand{\M}{\mathcal{M}}

\renewcommand{\b}{{\bm{b}}}
\newcommand{\rkb}{{\mathbf{r}_{\bm{b}}}}
\renewcommand{\a}{{\bm{a}}}
\newcommand{\rka}{{\mathbf{r}_{\bm{a}}}}

\newcommand{\y}{{\bm{y}}}
\newcommand{\haty}{{\hat{\bm{y}}}}

\def\eg{\emph{e.g.}} 
\def\ie{\emph{i.e.}} 
 
\def\etc{\emph{etc.}} 
\def\wrt{{w.r.t.\ }}

\title{Relational Surrogate Loss Learning}

\author{Tao Huang${}^1{}^2$, Zekang Li$^3$, Hua Lu$^4$, Yong Shan$^3$, Shusheng Yang$^4$, \\
\textbf{Yang Feng$^3$, Fei Wang$^5$, Shan You$^2$\thanks{Correspondence to: Shan You~$<$\texttt{youshan@sensetime.com}$>$.}, Chang Xu$^1$} \\
$^1$School of Computer Science, Faculty of Engineering, The University of Sydney\\
$^2$SenseTime Research\\
$^3$Key Laboratory of Intelligent Information Processing, \\
\ \ Institute of Computing Technology, Chinese Academy of Sciences (ICT/CAS)\\
$^4$Huazhong University of Science and Technology\\
$^5$University of Science and Technology of China
}

\iclrfinalcopy 
\begin{document}

\maketitle

\begin{abstract}
Evaluation metrics in machine learning are often hardly taken as loss functions, as they could be non-differentiable and non-decomposable, \eg, average precision and F1 score. This paper aims to address this problem by revisiting the surrogate loss learning, where a deep neural network is employed to approximate the evaluation metrics. Instead of pursuing an exact recovery of the evaluation metric through a deep neural network, we are reminded of the purpose of the existence of these evaluation metrics, which is to distinguish whether one model is better or worse than another. In this paper, we show that directly maintaining the relation of models between surrogate losses and metrics suffices, and propose a rank correlation-based optimization method to maximize this relation and learn surrogate losses. Compared to previous works, our method is much easier to optimize and enjoys significant efficiency and performance gains. Extensive experiments show that our method achieves improvements on various tasks including image classification and neural machine translation, and even outperforms state-of-the-art methods on human pose estimation and machine reading comprehension tasks. Code is available at: \url{https://github.com/hunto/ReLoss}.
\end{abstract}

\section{Introduction}
Evaluation metrics matter in machine learning since it depicts how well we want the models to perform. Nevertheless, most of them are non-differentiable and non-decomposable, thus we can not directly optimize them during training but resort to loss functions (or surrogate losses), which serve exactly as a proxy of task metrics. For example, pose estimation task uses percentage of correct keypoints (PCK)~\citep{yang2012articulated} to validate point-wise prediction accuracy, but it often adopts mean square error (MSE) as loss function. Neural machine translation task takes the sentence-level metric BLEU \citep{DBLP:conf/acl/PapineniRWZ02} to evaluate the quality of predicted sentences, while using word-level cross-entropy loss (CE Loss) in training.

Besides this manual proxy, some works~\citep{grabocka2019learning, patel2020learning} propose to learn surrogate losses which approximate the metrics using deep neural networks (DNN), so the optimization of metrics can be relaxed to a differentiable space. For example, taking predictions and labels as input, ~\citep{grabocka2019learning} approximates the outputs of surrogate losses and evaluation metrics by minimizing their L2 distances. Moreover, recent work even involves the prediction networks into the surrogate loss learning by alternatively updating the loss and predictions, \ie, they train the surrogate losses after every epoch during training, then use the latest optimized losses to train prediction networks in the next epoch. For instance, \citep{grabocka2019learning} mainly focuses on the simple binary classification while LS-ED~\citep{ patel2020learning} chooses to adopt the surrogate losses in the post-tuning stage (fine-tuning the models learned by original losses) and achieves promising improvements. However, these methods often suffer from heavy computational consumption and do not perform well on large-scale challenging datasets.

\begin{figure}[t]
	\centering
	\subfigure[Spearman's rank correlations] 
	{\includegraphics[height=0.25\textwidth]{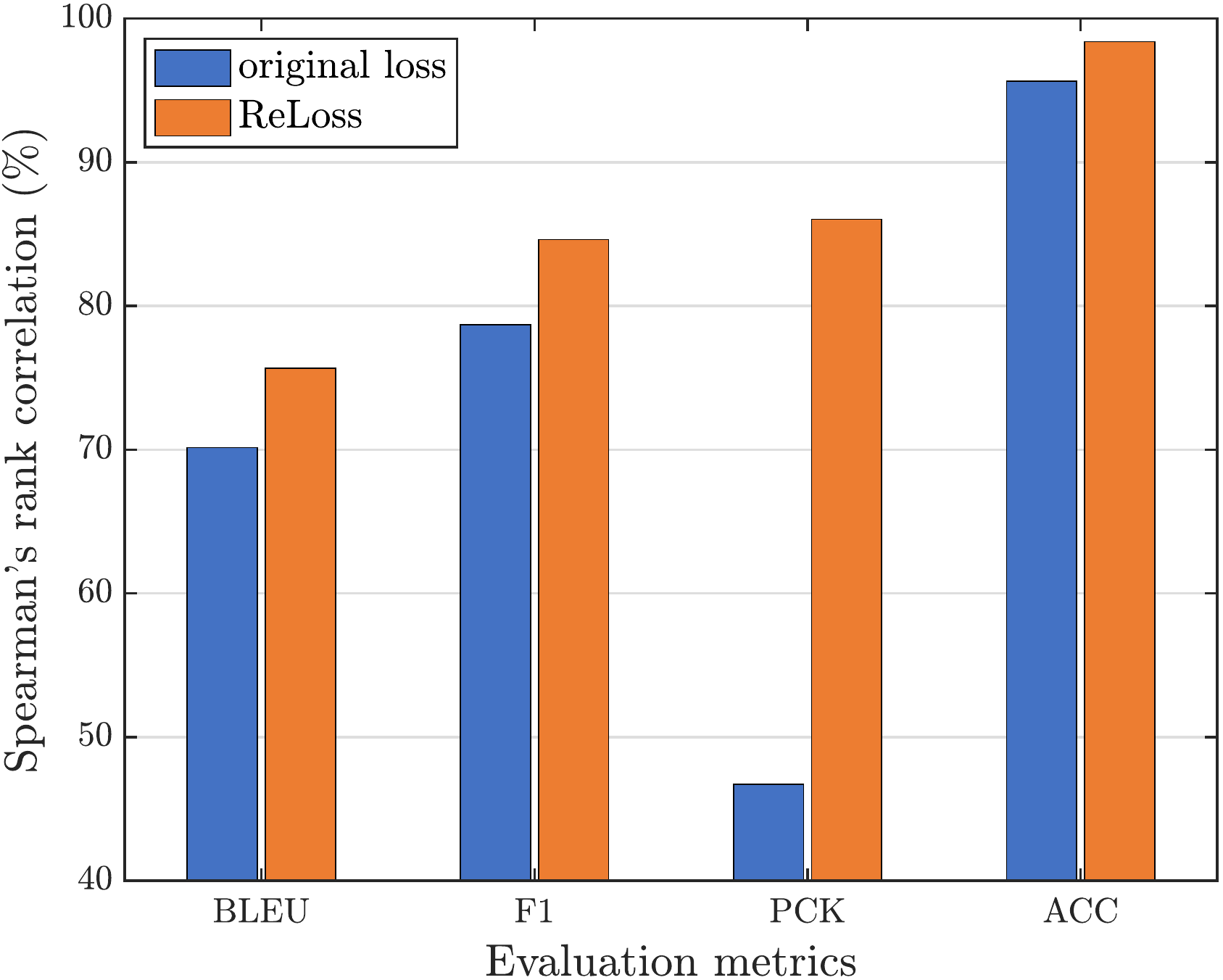}} 
	\hspace{1mm}
	\subfigure[Example on neural machine translation task]
	{\includegraphics[height=0.25\textwidth]{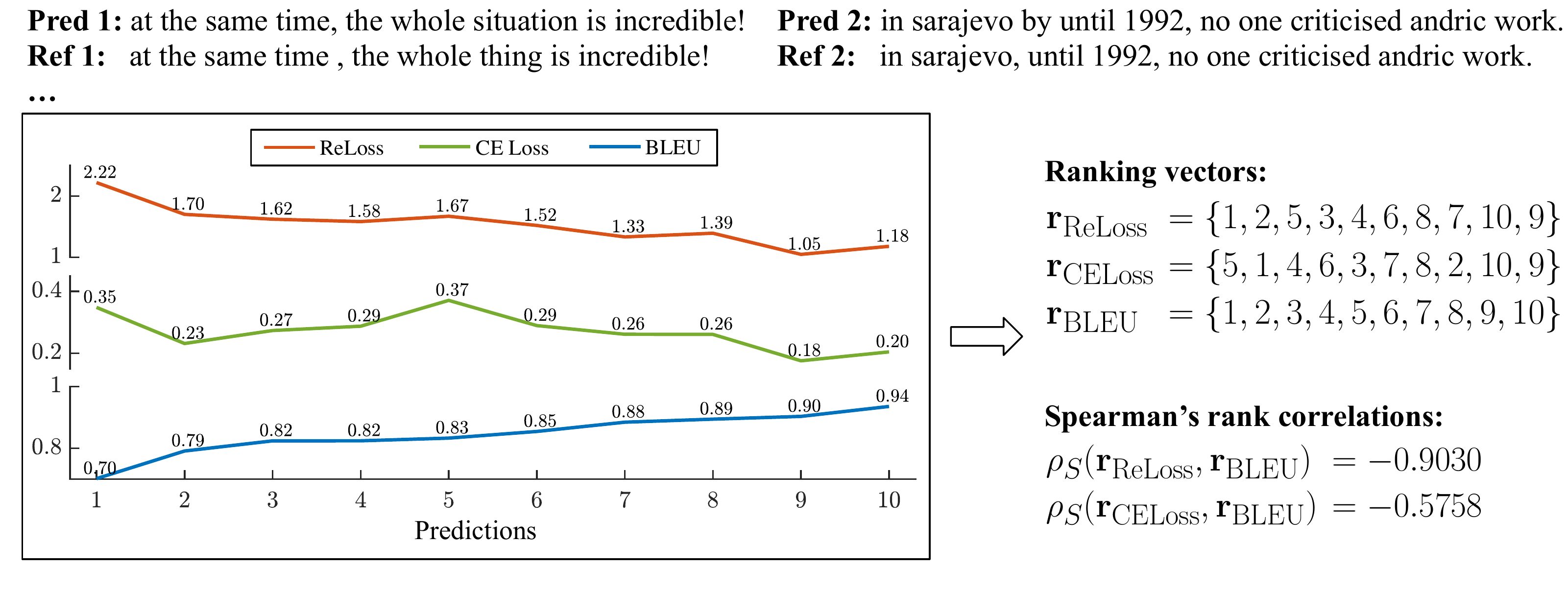}}
	\vspace{-4mm}
	\caption{(a) Our proposed ReLoss significantly improves the ranking correlations between losses and metrics on various tasks. (b) Taking neural machine translation task as an example, we sample 10 sentences from WMT16 RO-EN dataset, then measure the BLEU, cross entropy (CE) loss, and ReLoss with trained network and ground-truth references. Compared to the original CE loss, our ReLoss obtains a stronger rank correlation.}
	\vspace{-6mm}
	\label{fig:fig1}
\end{figure}

Both manual and learned surrogate losses follow an \textit{exact recovery} manner; namely, the surrogate losses should approximate the target metrics rigorously, and optimizing the surrogate loss is supposed to improve the evaluation metrics accordingly. However, this assumption does not always hold due to the approximation gap, bringing bias to the optimization and leading to sub-optimal results. Instead of pursuing an exact recovery of the evaluation metric, we are reminded of the purpose of metrics, which is to distinguish the performance of models. If a model has a smaller loss than the other model, its metric ought to be better. Nevertheless, current surrogate losses usually have weak relation with the evaluation metrics (\eg, CE Loss \& BLEU in Figure~\ref{fig:fig1} (b)). Ideally, the surrogate loss should maintain strong relation of evaluation metric to all models. 

In this paper, we leverage the ranking correlation as the relation between surrogate losses and evaluation metrics. Then a natural question raises, \textit{if the loss functions only require accurate relative rankings to discriminate the models, why do we need to approximate the metrics exactly? } In this way, we propose a method named Relational Surrogate Loss (ReLoss) to maximize this rank correlation directly. Concretely, our ReLoss directly leverages the simple Spearman's rank correlation~\citep{dodge2008concise} as the learning objective. By adopting differentiable ranking method, the ranking correlation coefficient can be maximized through gradient descent. Compared to exactly recovering the metrics, our correlation-based optimization is much easier to learn, and our ReLoss, which is simply constructed by multi-layer perceptions, aligns well with the metrics and obtains significantly better correlations compared to the original losses. For example, the commonly used loss MSE in pose estimation only has $46.71\%$ Spearman's rank correlation coefficient with the evaluation metric PCK, while our ReLoss enjoys $84.72\%$ relative improvement (see Table~\ref{tab:correlation} and Figure~\ref{fig:fig1}).

Our ReLoss generalizes well to various tasks and datasets. We learn ReLoss using randomly generated data and pre-collected network outputs, then the learned losses are integrated into the training of prediction networks as normal loss functions (\eg, cross-entropy loss), without any further fine-tuning. Note that we use the same surrogate losses with the same weights in each task, and we find that it is sufficient to obtain higher performance. Compared to previous works, our method is much easier to optimize and enjoys significant efficiency and performance improvements. Extensive experiments on the synthetic dataset and large-scale challenging datasets demonstrate our effectiveness. Moreover, our method outperforms the state-of-the-art methods in human pose estimation and machine reading comprehension tasks. For example, on human pose estimation task, our ReLoss outperforms the state-of-the-art method DARK~\citep{zhang2020distribution} by $0.2\%$ on COCO test-dev set; on machine reading comprehension task, we achieve new state-of-the-art performance on DuReader 2.0 test set, outperforming all the competitive methods, and even obtain $7.5\%$ better ROUGE-L compared to human performance.

\section{Related Work}
\textbf{Surrogate loss learning.} Since most of the metrics in deep learning tasks are non-differentiable and non-decomposable (\eg, accuracy, F1, AUC, AP, \etc), surrogate losses aim to approximate the metrics to make them differentiable using neural networks.
\citep{grabocka2019learning} first proposes to learn surrogate losses by approximating the metrics of tasks through a neural network, and the losses are optimized jointly with the prediction model via bilevel optimization. \citep{patel2020learning} learns the surrogate losses via a deep embedding where the Euclidean distance between the prediction and ground truth corresponds to the value of the metric. However, it is hard to obtain a precise prediction by directly optimizing the surrogate loss with such a strong constraint. We remind that the role of loss functions is to determine which model is better, but with the unavoidable existence of approximation gap, this determinability does not always hold. In addition, these methods both train the surrogate losses alternately with prediction networks, resulting in noticeable efficiency and generability deduction compared to regular losses. In our paper, instead of only focusing on point-to-point recovery, which ignores the rankings between relative values of metrics, we ease the optimization constraint by explicitly learning our ReLoss with rank correlation, and enjoy significant performance and efficiency improvements.

\textbf{Differentiable sorting \& ranking.} Differentiable sorting and ranking algorithms~\citep{adams2011ranking, grover2018stochastic, blondel2020fast, petersen2021differentiable} can be used in training neural networks with sorting and ranking supervision. Recent approach \citep{blondel2020fast} proposes to construct differentiable sorting and ranking operators as projections onto the permutahedron, \ie, the convex hull of permutations, and using a reduction to isotonic optimization. \citep{petersen2021differentiable} proposes differentiable sorting networks by relaxing their pairwise conditional swap operations. In this paper, we can use any of these differentiable ranking algorithms to generate differentiable ranking vectors, then directly optimize the rank correlation coefficient for the supervision of our surrogate losses. The algorithm in \citep{petersen2021differentiable} is adopted for better performance.

\begin{figure}[t]
	\centering
	\subfigure[] 
	{\includegraphics[height=0.22\textwidth]{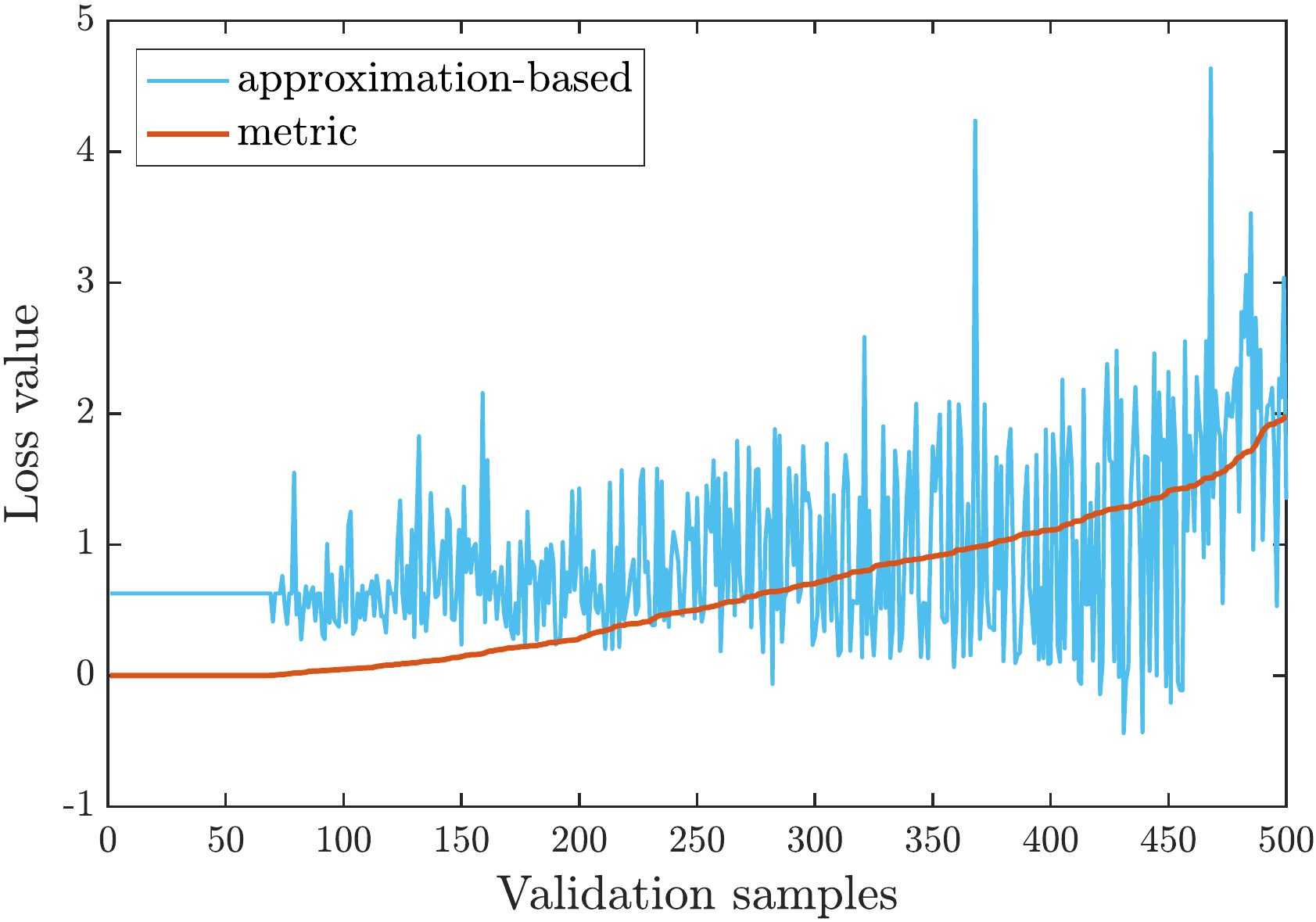}} 
	\hspace{-0mm}
	\subfigure[]
	{\includegraphics[height=0.22\textwidth]{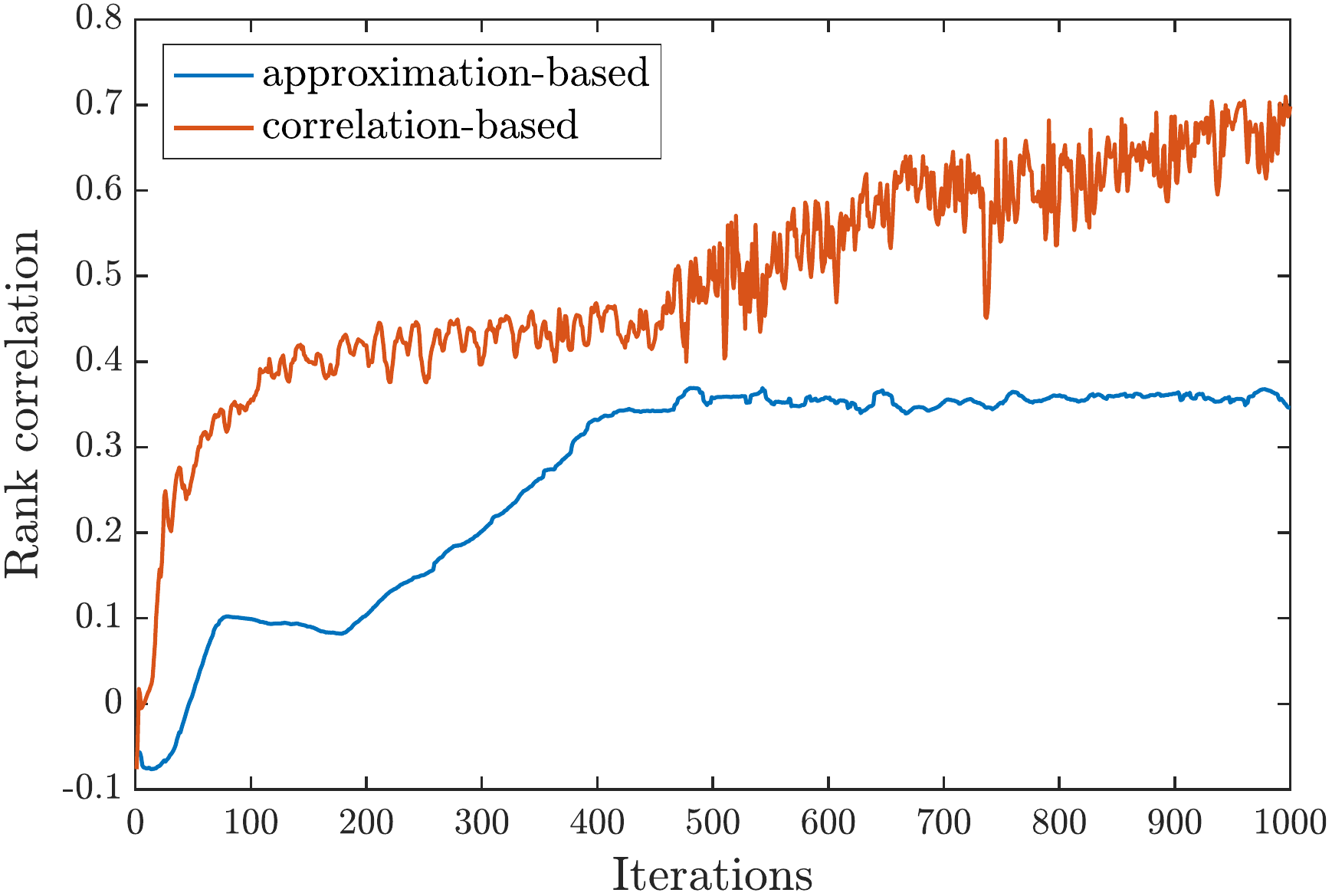}}
	\hspace{-0mm}
	\subfigure[]
	{\includegraphics[height=0.22\textwidth]{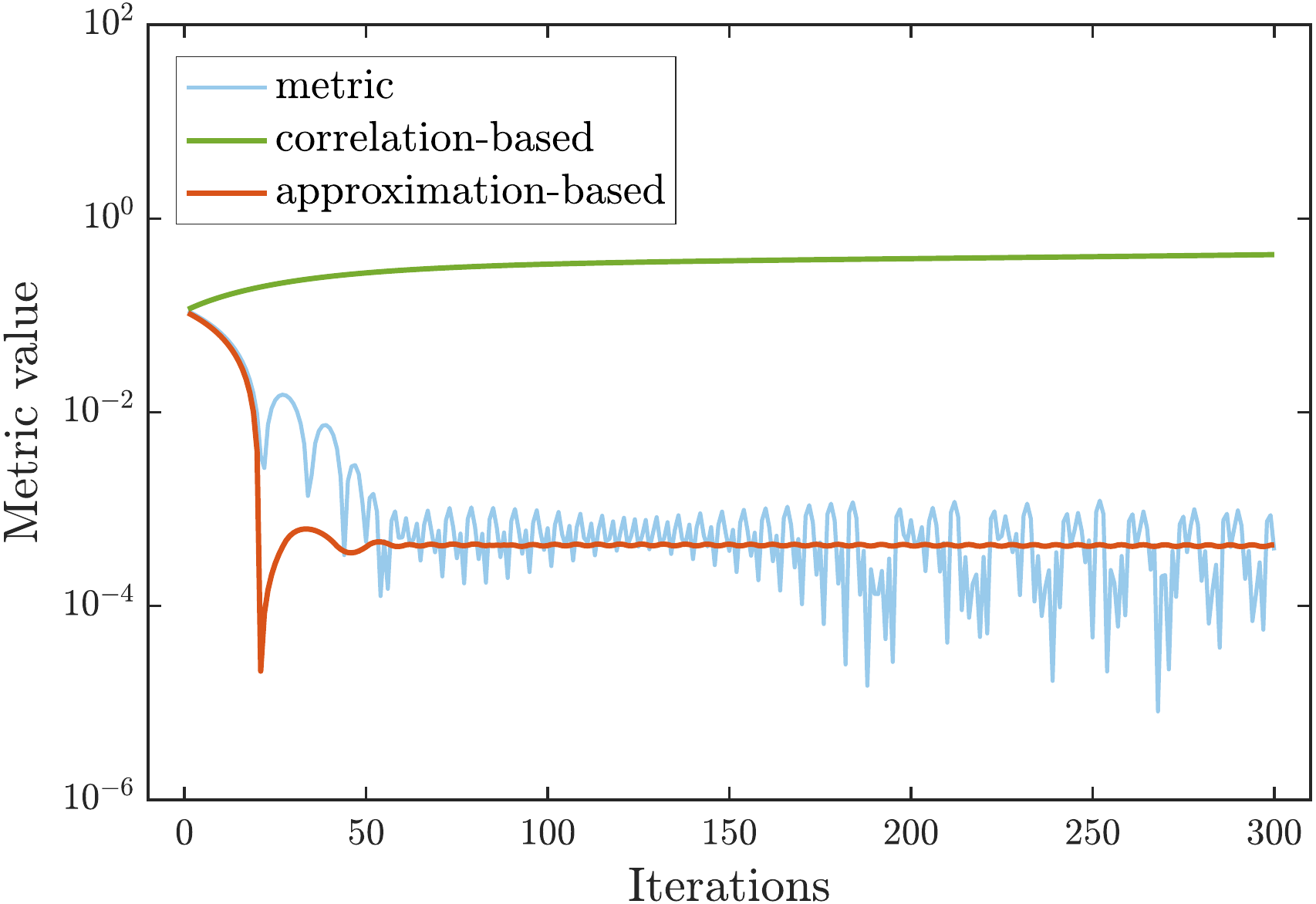}}
	\vspace{-4mm}
	\caption{Visualization of our toy experiments on the synthetic dataset. (a) Visualization of outputs of approximation-based surrogate loss and evaluation metric on the validation set. (b) Curves of Spearmans' rank correlations between surrogate losses and evaluation metric in the training of losses. (c) Evaluation curves of different losses in training, lower is better.}
	\vspace{-2mm}
	\label{fig:synthetic}
\end{figure}

\section{Preliminaries}

For a given task with a metric function $\M(\y, \haty)$, where $\y$ and $\haty$ denote the predicted labels and ground-truth labels, respectively, its loss function $\L(\y, \haty)$ can be formulated as:
\begin{equation}
   \L(\y, \haty) = f(\y, \haty) ,
\end{equation}
where $f$ can be any function with output $\in \mathbb{R}^1$.

In this paper, we tend to use a learned DNN ($f_\textrm{DNN}$) with weights $\bm{\theta}_l$ as a surrogate loss, i.e., 
\begin{equation}
   \L(\y, \haty; \bm{\theta}_l) = f_\textrm{DNN}(\y, \haty; \bm{\theta}_l) .
\end{equation}

The surrogate losses are learned with the networks' outputs $\y$ and the corresponding metric values $\M(\y, \haty)$, \ie,
\begin{equation}
    \bm{\theta}_l^* = \argmin_{\bm{\theta}_l} \O_{\textrm{s}} (\L(\y, \haty; \bm{\theta}_l), \M(\y, \haty)) ,
\end{equation}
where $\O_{\textrm{s}}$ is the learning objective of surrogate loss. The prediction networks with weights $\bm{\theta}_{m}$ are then optimized by descending the learned surrogate losses $\L(\y, \haty; \bm{\theta}_l^*)$, \ie, 
\begin{equation}
    \bm{\theta}_{m}^* = \argmin_{\bm{\theta}_{m}} \L(\y, \haty; \bm{\theta}_l^*) .
\end{equation}

\textbf{Approximation-based optimization.}\ \ To learn a surrogate loss \wrt a metric, an intuitive idea is to approximate the metric's outputs, \ie,
learn the surrogate losses by minimizing the distances between the outputs of surrogate losses and their corresponding metric values, which is adopted in previous works~\citep{grabocka2019learning, patel2020learning},
their learning objective $\O_{\textrm{s}}$ is 
\begin{equation}
    \O_{\textrm{s}}(\L(\y, \haty; \bm{\theta}_l), \M(\y, \haty)) = \parallel \L(\y, \haty; \bm{\theta}_l) - \M(\y, \haty) \parallel_2^2 ,
\end{equation}
we call this optimization as approximation-based optimization.

However, it is hard for a DNN to fully recover the evaluation metric. We conduct toy experiments using a random weighted DNN with output $\in\mathbb{R}^1$ as an evaluation metric, and then the surrogate loss is learned using limited observations of the metric. As illustrated in Figure~\ref{fig:synthetic} (a), since approximating a random network with random inputs is challenging, the errors between surrogate loss learned by approximation-based optimization and metric values are noticeably large. In order to validate the effectiveness of losses in training, we then train the input data with metric or learned losses, as shown in Figure~\ref{fig:synthetic} (c), we illustrate the curves of metric values \wrt learned input data during training, and directly using metric as loss function obtains best metric value (lower is better), but the performance of input data using approximation-based loss is getting worse.

\section{Learning Relational Surrogate Loss}

\subsection{Relation as Rank Correlation}

Based on the previous discussion, the prior works adopt an unnecessary constraint by enforcing the surrogate losses to fully recover the evaluation metrics. However, the loss function only needs to have the same ranking relation to the metrics, \ie, we just need to make the surrogate losses have the same ranking as metrics. In this paper, we obtain the relation between surrogate losses and evaluation metrics by using rank correlation as the learning objective, which we call correlation-based optimization. 


The relation between surrogate losses and evaluation metrics is measured by ranking correlation, which is a statistic that measures the relationship between rankings of the same variable. A ranking correlation coefficient measures the degree of similarity between two rankings and can be used to assess the relation's significance. If the surrogate loss fully correlates to the evaluation metric, the descent of loss value will always obtain better metric values.

\textbf{Spearman's rank correlation.}\ \ 
For optimization of surrogate losses, we use the most commonly used Spearman's rank correlation~\citep{dodge2008concise}. For two vectors $\a$ and $\b$ with size $n$, the Spearman's rank correlation is defined as:
\begin{equation} \label{eq:spearman}
   \rho_S (\a, \b) = \frac{\textrm{Cov}(\rka, \rkb)}{\textrm{Std}(\rka) \textrm{Std}(\rkb)} = \frac{\frac{1}{n-1}\sum_{i=1}^n(\rka_i - E(\rka))(\rkb_i - E(\rkb))}{\textrm{Std}(\rka) \textrm{Std}(\rkb)},
\end{equation}
where $\rka$ is the rank vector of $\bm{a}$, $\textrm{Cov}(\rka, \rkb)$ is the covariance of the rank vectors, $\textrm{Std}(\rka)$ denotes the standard derivation of $\rka$. 

\subsection{Learning Losses by Maximizing Rank Correlation}
\textbf{Correlation-based optimization.}\ \ We use Spearman's rank correlation as the objective to learn our surrogate losses, since the loss should have a negative correlation \wrt the metric (higher is better), our objective is to minimize the Spearman's rank correlation coefficient, \ie,
\begin{equation} \label{eq:obj_correlation}
    \O_{\textrm{s}}(\L(\y, \haty; \bm{\theta}_l), \M(\y, \haty)) = \rho_S(\L(\y, \haty; \bm{\theta}_l), \M(\y, \haty))
\end{equation}

Since the computation of rank vectors $\rka$ and $\rkb$ in Eq.(\ref{eq:spearman}) is not differentiable, we adopt one of the differentiable ranking methods \citep{petersen2021differentiable} to obtain differentiable ranking vectors, and empirically find that the errors in differentiable approximation is negligible and our learned correlation can be very close to the optimal value, \ie, $\O_{\textrm{s}} = -1$.

As shown in Figure~\ref{fig:synthetic}, compared to approximation-based optimization, the surrogate loss learned by our correlation-based optimization obtains higher rank correlation and faster convergent speed. Besides, optimizing with our correlation-based loss achieves significantly better performance than approximation-based optimization, and is more stable than the original loss (evaluation metric).

\textbf{Learning with Gradient Penalty.}\ \ 
In our paper, we can directly backward through the surrogate loss to obtain its gradients. However, we find the first-order derivative of the learned ReLoss w.r.t. the prediction $\y$ changes rapidly since we only constrain the correlation in Eq.(3), which result in either vanishing or exploding gradients. Nevertheless, in the optimization of networks, we want a loss with smooth gradients to train the networks steadily. 

Following~\citep{gulrajani2017improved, patel2020learning}, we now propose an alternative way to smooth the gradients by enforcing the Lipschitz constraint. A differentiable function is 1-Lipschitz if and only if it has gradients with norm at most 1 everywhere, so we consider directly constraining the gradient norm of the loss’s output \wrt its input, \ie
\begin{equation}
   \L_{\textrm{penalty}} = (\parallel \nabla_y \L(\y, \haty; \bm{\theta}_l) \parallel_2 - 1)^2 .
\end{equation}

This penalty of gradients has been shown to enhance the training stability for generative adversarial networks~\citep{gulrajani2017improved}. Our objective of surrogate loss learning in Eq.(\ref{eq:obj_correlation}) becomes
\begin{equation} \label{eq:obj_correlation_with_penalty}
    \O_{\textrm{s}}(\L(\y, \haty; \bm{\theta}_l), \M(\y, \haty)) = \rho_S(\L(\y, \haty; \bm{\theta}_l), \M(\y, \haty)) + \lambda \L_{\textrm{penalty}} ,
\end{equation}
we use $\lambda=10$ in our experiments.

\subsection{Pipeline} \label{sec:pipeline}

Now we illustrate how to learn our ReLoss. Different from previous works~\citep{grabocka2019learning, patel2020learning} which train the surrogate loss and prediction network alternatively as bilevel optimization, we want our surrogate loss to be general as vanilla loss (\eg, cross-entropy loss). Since we learn the surrogate loss with a much weaker constraint, our surrogate loss can generalize better to the whole distribution of outputs and metric values to train the surrogate loss once for all, without further fine-tuning.

Our training strategy of surrogate loss is summarized in Algorithm~\ref{alg:train_loss}. The training data of surrogate losses is the combination of randomly generated data $G_R$ and the outputs of models $G_M$. Concretely, we design a random generator to produce random outputs and labels uniformly for $G_R$, while for $G_M$, we use the intermediate checkpoints of the prediction networks trained by original loss to predict the outputs of train data. Each batch of training data is generated from $G_R$ or $G_M$ with probabilities $p$ and $1 - p$, respectively.

\input{alg_train_loss.tex}

\textbf{Usage of learned ReLoss.} The learned ReLoss can be fixed and then integrated into the training of prediction networks, \ie, we only change the loss function in training, without any modification on training strategy, network architecture, \etc Besides, we emprically find that ReLoss would achieve better performance if combined with the regular loss. In this case, the regular loss might act as a regularization term, and bring a decent prior for the ReLoss to enhance the optimization.   

\section{Experiments}

To fully experiment with the effectiveness and generability of our ReLoss, we conduct experiments on both computer vision and natural language processing tasks. In computer vision, we experiment on image classification and human pose estimation tasks, while in natural language processing, we experiment on machine reading comprehension and neural machine translation tasks.

We first show the rank correlations of original losses and our learned surrogate losses to the metrics in Table~\ref{tab:correlation}. Spearman's~\citep{dodge2008concise} and Kendall's Tau~\citep{kendall1938new} are two commonly used coefficients to measure the ranking correlations between two vectors. We can see that in all our experimented tasks, our ReLoss achieves higher correlations compared to the original losses. Notably, even for cross-entropy (CE) loss, which has been shown to align well with the misclassification rate, our surrogate loss still performs better on classification tasks with the metric accuracy (ACC).

\input{tables/table_correlation.tex}

\subsection{Computer Vision}

\textbf{Image classification.}\ \ 
We conduct experiments on three benchmark datasets CIFAR-10, CIFAR-100~\citep{krizhevsky2009learning}, and ImageNet~\citep{Imagenet}. On CIFAR-10 and CIFAR-100 datasets, we train ResNet-56~\citep{he2016deep} with original CE loss and our surrogate loss and report their accuracies on the test set with mean and standard derivation of $5$ runs. While on ImageNet dataset, we train ResNet-50 and MobileNet V2~\citep{sandler2018mobilenetv2}, their accuracies on validation set are reported. Notably, all experiments use the same surrogate loss with the same weights.

Table~\ref{tab:image_cls} shows the evaluation results. We can see that, though the original CE loss obtains a very high correlation ($\sim0.96$ in Table~\ref{tab:correlation}), by integrating our surrogate loss with higher correlation, the performance can still be improved. Note that we use the same surrogate loss with fixed weights in these three datasets, which means that our loss can generalize to different image classification datasets and gain the improvements with negligible additional cost.

\input{tables/table_image_cls.tex}

\textbf{Human pose estimation.}\ \ 
Human pose estimation (HPE) aims to locate the human body and build body skeleton from images. It is difficult to precisely evaluate the performance of HPE since many features need to be considered (\eg, the quality of body parts, the precision of each keypoints). As a result, many metrics are proposed for HPE. Percentage of correct keypoints (PCK)~\citep{yang2012articulated} and Average Precision (AP) are two of the most commonly used ones. However, current methods usually adopt mean square error (MSE) to minimize the distance between predicted heatmap and target heatmap, which correlates weakly with the evaluation metrics. 

In our experiments, we choose to approximate PCK@0.05 since it better reflects the quality of each keypoint, and our ReLoss achieves significant improvement on rank correlation compared to the original MSE loss. We use the most widely used large-scale dataset COCO~\citep{lin2014microsoft} to evaluate our performance, and the results are summarized in Table~\ref{tab:pose}. We can see that, on validation set, our ReLoss significantly improves the baseline methods, and the AP$^{75}$ improves the most since our ReLoss aligns PCK for better keypoint localization. On test-dev set, we integrate our ReLoss into state-of-the-art method DARK~\citep{zhang2020distribution} and achieve improvements on all the metrics.

\input{tables/table_pose_sota.tex}

\subsection{Natural Language Processing}
The gaps between loss functions and evaluation metrics on natural language processing tasks are severer since the tasks often use sentence-level evaluation metrics (\eg, BLEU and ROUGE-L) but adopt word-level cross-entropy loss in training.

\textbf{Machine reading comprehension.}\ \ The task of machine reading comprehension (MRC) aims to empower machines to answer questions after reading articles. Concretely, with a given question, the models are required to locate a segment of text from the corresponding reading passage, which is most probably the answer. We use F1 score as the evaluation metric to learn surrogate loss, and experiment on two typical MRC datasets SQuAD~\citep{rajpurkar2016squad} and DuReader
\citep{he2018dureader}. SQuAD evaluates performance using F1-score, and DuReader uses ROUGE-L~\citep{lin2004rouge} and BLEU-4~\citep{DBLP:conf/acl/PapineniRWZ02}.

The evaluation results are summarized in Table~\ref{tab:mrc} and Table~\ref{tab:mrc_squad}. On DuReader 2.0 dataset, our ReLoss gains improvements on dev set, and achieves state-of-the-art performance on test set. On SQuAD 1.1 dataset, we also achieve improvements compared to the baseline method.

\input{tables/table_mrc_sota.tex}

\textbf{Neural machine translation.}\ \ 
Neural machine translation (NMT) aims to translate a sentence from the source to the target language with an end-to-end neural model. The evaluation metric of NMT is BLEU~\citep{DBLP:conf/acl/PapineniRWZ02}, which measures the n-gram overlap between the generated translation and the reference. 
We conduct experiments on the Non-Autoregressive neural machine Translation (NAT) task, in which the model generates target words independently and simultaneously. Since the output of NAT cannot be properly evaluated through word-level cross-entropy loss due to the multimodality problem in language, the correlation between cross-entropy loss and translation quality is weak, limiting the NAT performance.

Table \ref{tab:nat} shows the NAT evaluation results on WMT-16 EN$\rightarrow$RO and RO$\rightarrow$EN datasets. We conduct experiments based on the NAT-base and a strong baseline BoN-$L_1$ (N=2) \citep{DBLP:journals/corr/abs-2106-08122}, which introduce the BoN loss to fine tune NAT-base by modeling the bag of ngrams in the sentence. We integrate ReLoss into the baseline methods NAT-base ~\citep{gulrajani2017improved} and BoN-$L_1$ (N=2), and the evaluation results show that our ReLoss can improve both of them.
\input{tables/table_nat.tex}

\subsection{Ablation Studies}
\textbf{Compare with LS-ED.}\ \ Prior work LS-ED~\citep{patel2020learning} aims to post-tune the scene text recognition (STR) model using a surrogate loss, which is learned with approximation-based optimization. In order to compare our method with LS-ED, we conduct experiments on the same settings. Following LS-ED, we learn the surrogate loss using edit distance, then fine-tune the trained model using our learned loss (without using original loss for fair comparisons). The results in Table~\ref{tab:str} show that our ReLoss significantly outperforms the baselines CE and LS-ED. Note that we only train our ReLoss once then integrate it into training, indicating that our loss is more efficient and general. 

\input{tables/table_str.tex}

\textbf{Transferability of learned ReLoss.}\ \ 
In all our experiments, we use the same surrogate loss in each task. If we learn different surrogate losses on specific datasets, would the performance be better? To validate this, we conduct experiments to train ReLoss independently on each dataset, as shown in Table~\ref{tab:ab_trans}. The ReLoss transferred from ImageNet dataset performs similar to the consistent ReLoss learned on corresponding datasets. It might be because we train the ReLoss using predicted and randomly generated data, and it is sufficient to cover different distributions of datasets on image classification. 
We also experiment on NMT task. As shown in Table \ref{tab:nat}, we train ReLoss on both EN$\rightarrow$RO and RO$\rightarrow$EN, and the results using either of them to train the networks are similar, which demonstrates that the learned ReLoss is language-independent and can bring similar improvements on the other translation direction.

\textbf{Comparison of approximation-based and our correlation-based optimization.}
In Figure~\ref{fig:synthetic}, we compare our ReLoss with approximation-based methods on the synthetic dataset. Now we further conduct experiments on image classification task to show our superiority. Concretely, we learn the surrogate losses with the same architecture using approximation-based or our correlation-based optimization, then integrate them to train networks on CIFAR datasets. As shown in Table~\ref{tab:ab_opt}, our ReLoss with correlation-based optimization obtains the highest accuracies compared to the CE loss and approximation-based loss. Note that the standard derivations of accuracies of approximation-based loss are much larger than CE loss and correlation-based loss; this might be because the imprecise rankings and gradients in approximation-based loss weaken the training stability.

\input{tables/table_ab_opt.tex}

\textbf{Integrating ReLoss with / without regular losses.}\ \ We empirically find that the prediction networks using our ReLoss converge very fast at the beginning of training, then the performance will increase very slowly or even get worse, the experiments on synthetic dataset show the similar trend (see Figure~\ref{fig:synthetic} (c)). A possible reason is that there exist some data points that surrogate losses can not predict accurately, making the optimization fall into local minima. For better performance, we use the regular loss as a regularization term to help the surrogate losses jump out local minima. We conduct experiments to show the differences by integrating ReLoss with or without regular loss. As summarized in Table~\ref{tab:ab_regular}, the performance drops if not adding regular loss in training, showing that the regular losses can bring a decent prior for ReLoss to achieve better performance.

\input{tables/table_ab_trans_single.tex}

\subsection{Complexity Analysis}
Denoting the training iterations of surrogate losses as $T_l$, the training epochs and iterations in each epoch of prediction networks are $E_m$ and $T_m$, respectively, the runtime complexity of our ReLoss is $\O(T_l + E_m \times T_m)$. For comparison, the runtime complexity of regular loss is $\O(E_m \times T_m)$, while for previous surrogate loss learning method~\citep{grabocka2019learning}, it trains the surrogate losses after every iteration and has a runtime complexity of $\O(E_m \times T_l + E_m \times T_m)$.
Note that our additional cost $\O(T_l)$ of learning ReLoss costs only 0.5 GPU hour on image classification with a single NVIDIA TITAN Xp GPU, and we only need to train ReLoss once for each task, reducing much computational cost compared to previous works.

\section{Conclusion}
As a proxy of the evaluation metric, loss function matters in machine learning since it controls the optimization of networks. However, it is often hard to design a loss function with strong relation to the evaluation metric. In this paper, we aim to address this problem by learning surrogate losses using deep neural networks. Unlike previous works that pursue an exact recovery of the evaluation metric, we are reminded of the essence of the loss function and evaluation metric, which is to distinguish the performance of models, and show that directly maximizing the rank correlation between surrogate loss and evaluation metric can learn better loss. How to design and learn a more robust and general surrogate loss would be a valuable aspect to improve this work.

\nocite{you2020greedynas}
\nocite{you2017learning}
\nocite{su2020locally}

\section*{Acknoledgements}
Chang Xu was supported by the Australian Research Council under Project DP210101859 and the University of Sydney SOAR Prize.

\bibliography{iclr2022_conference}
\bibliographystyle{iclr2022_conference}

\newpage
\appendix
\section{Appendix}

\subsection{Neural Architectures of Our Surrogate Losses}
\textbf{Image classification.}\ \ In order to make the learned surrogate loss generalize to all the classification tasks which take accuracy as the metric, we use the logits $\y_{\textrm{pos}}$ with positive labels as the input of our neural network, which is the same as cross-entropy loss, and the outputs of the surrogate loss are simply computed through 4-layer perceptions with intermediate activations, \ie, 
\begin{equation}
    l = \textrm{Mean}(\textrm{FC}(\textrm{ELU}(\textrm{FC}(\textrm{ELU}(\textrm{FC}(\textrm{ELU}(\textrm{FC}(\y_{\textrm{pos}})))))))) ,
\end{equation}
we use ELU activation~\citep{clevert2015fast} for stable gradients since it is $\textrm{C}^\infty$ continuous.

\textbf{Machine reading comprehension and neural machine translation.}\ \ Since the evaluation metrics of MRC and NMT are computed by a sequence of texts, based on the architecture in image classification, we use additional self-attention mechanisms~\citep{vaswani2017attention} to extract sequential information. 

\textbf{Human pose estimation.}\ \ Given the prediction heatmap and target heatmap, the original MSE loss is used to minimize the distance between these two heatmaps. Our ReLoss first embeds these two heatmaps into two hidden vectors, then computes the MSE loss between them as the final loss.

\subsection{Training Strategies}
\textbf{Surrogate loss.}\ \ We train the surrogate losses using Adam optimizer with a fixed learning rate of 0.01, and the weight decay is set to 1e-4.

\textbf{Image classification.}\ \ The reported models are trained using the same code, with the only difference in the loss function. On CIFAR-10 and CIFAR-100 datasets, we train ResNet-20 for 200 epochs with an initial learning rate of 0.1, which decays 0.1 at 100th and 150th epochs, the batch size is set to 128 with cutout~\citep{devries2017improved} data augmentation, we run each experiment 5 times with different random seeds and report their mean accuracy with standard derivation. On ImageNet, we follow the same training strategy as in torchvision\footnote{https://github.com/pytorch/vision/tree/main/references/classification}~\citep{marcel2010torchvision}. Concretely, we train ResNet-50 for 120 epochs with an initial learning rate of 0.1, a step learning rate scheduler which decays 0.1 every 30 epochs is adopted. While for MobileNet V2, we train it for 300 epochs with 4e-5 weight decay, a cosine learning rate scheduler is adopted with an initial learning rate of 0.045. The batch sizes for ResNet-50 and MobileNet V2 are both set to 32. We use SGD optimizer with 0.9 momentum on all datasets. Note that all the experiments use the same surrogate loss with the same weights.

\textbf{Human pose estimation.}\ \ We train ResNet-50, HRNet-W32, and DARK-HRNet-W48 following the default configurations in MMPose~\citep{mmpose2020}. Concretely, the models are trained with Adam optimizer for 210 epochs, and a step learning rate scheduler is adopted with initial value 5e-4, which decays 0.1 at 170th and 200th epochs. The total batch sizes of 8 GPUs with input size $256\times192$ and $384\times288$ are $512$ and $256$, respectively.

\textbf{Machine reading comprehension.}\ \ We train the networks using Adam optimizer with weight decay $0.01$, a linear learning rate strategy which warmups $0.1$ epoch and decays $2$ epochs is adopted. On DuReader 2.0 dataset, the batch size is set to $32$; we train MacBERT-base and MacBERT-large with learning rates 3e-5 and 2e-5, respectively. On SQuAD 1.1 dataset, we train BERT-base and BERT-large with batch sizes $32$ and $2$, and the learning rates are 5e-5 and 1e-5, respectively.

\textbf{Neural machine translation.} For WMT16 EN-RO, we use the WMT 2016 corpus, which consists of 610K sentence pairs for training. We take news-dev-2016 and news-test-2016 as development and test sets. We learn a joint BPE model with 32K operations and share the vocabulary for source and target languages.
As knowledge distillation \citep{DBLP:journals/corr/HintonVD15,DBLP:conf/emnlp/KimR16} has been proven to be crucial for training NAT models, we first train an auto-regressive transformer model \citep{vaswani2017attention}  as the teacher and then apply sequence-level knowledge distillation to construct the corpus for training NAT models. 
The NAT-base takes the same architecture as the base transformer model except that we modify the attention mask of the decoder for not masking the future tokens. We use a target length predictor to predict the length of the target sentence. We use golden length during the training and the predicted length during the inference.
For training ReLoss, we only use the outputs of the model.
For NAT-base, the number of training steps is 200K. We select the checkpoint based on the validation set. We add ReLoss with a factor of 1 on the CE loss to fine-tune the NAT-base for 10k steps with a batch size of 32 and a fixed learning rate of 1e-5. 
For BoN-L1 (N=2) \citep{DBLP:journals/corr/abs-2106-08122}, we reproduce the results using the public repo \footnote{https://github.com/ictnlp/Seq-NAT}. We combine ReLoss with the BoN Loss to fine-turn the model for 3K steps with a batch size of 512, which keeps the same with the BoN-L1 (N=2).

\subsection{More Ablation Studies}
\begin{figure}[h]
	\centering
	\subfigure[] 
	{\includegraphics[height=0.35\textwidth]{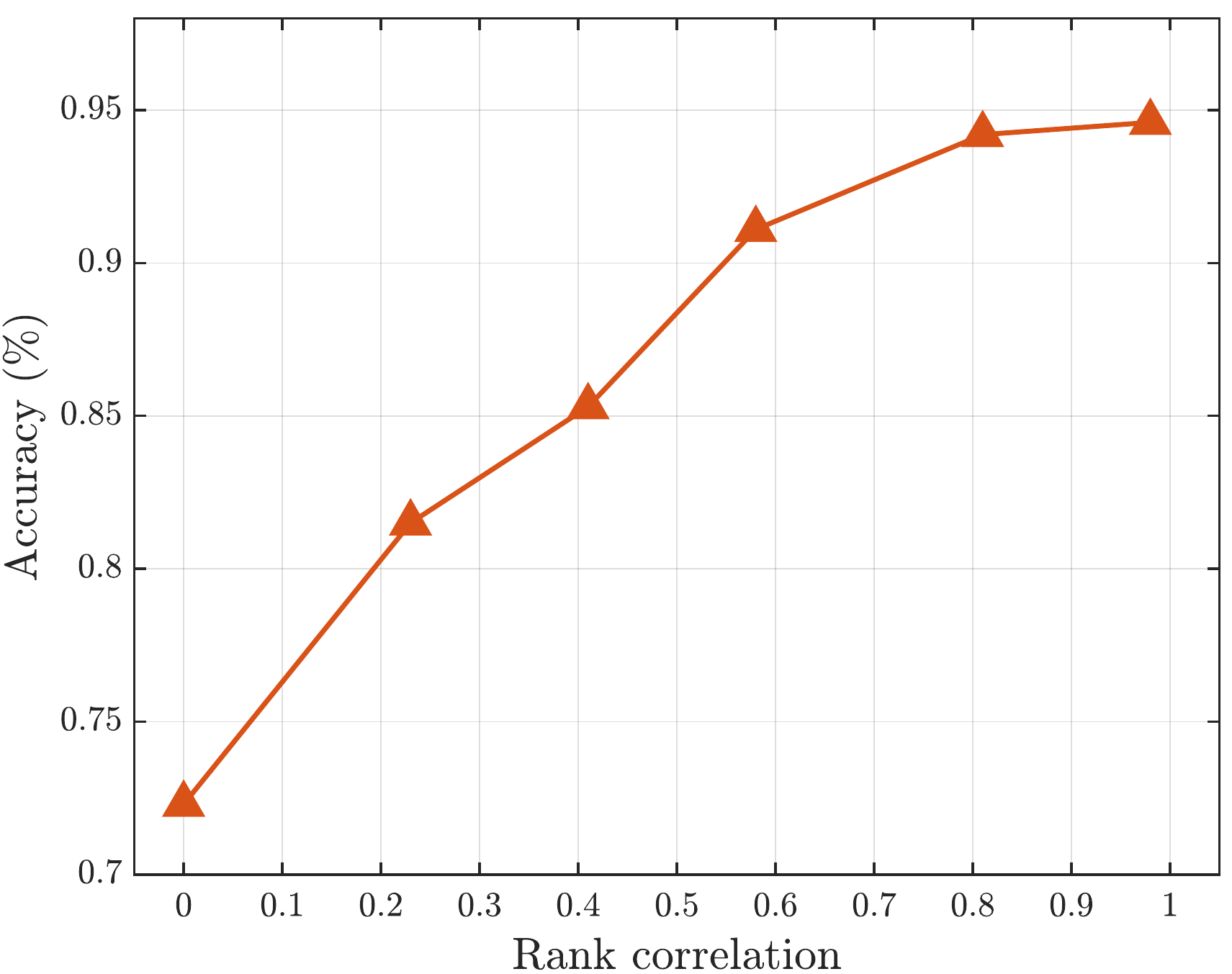}} 
	\subfigure[]
	{\includegraphics[height=0.35\textwidth]{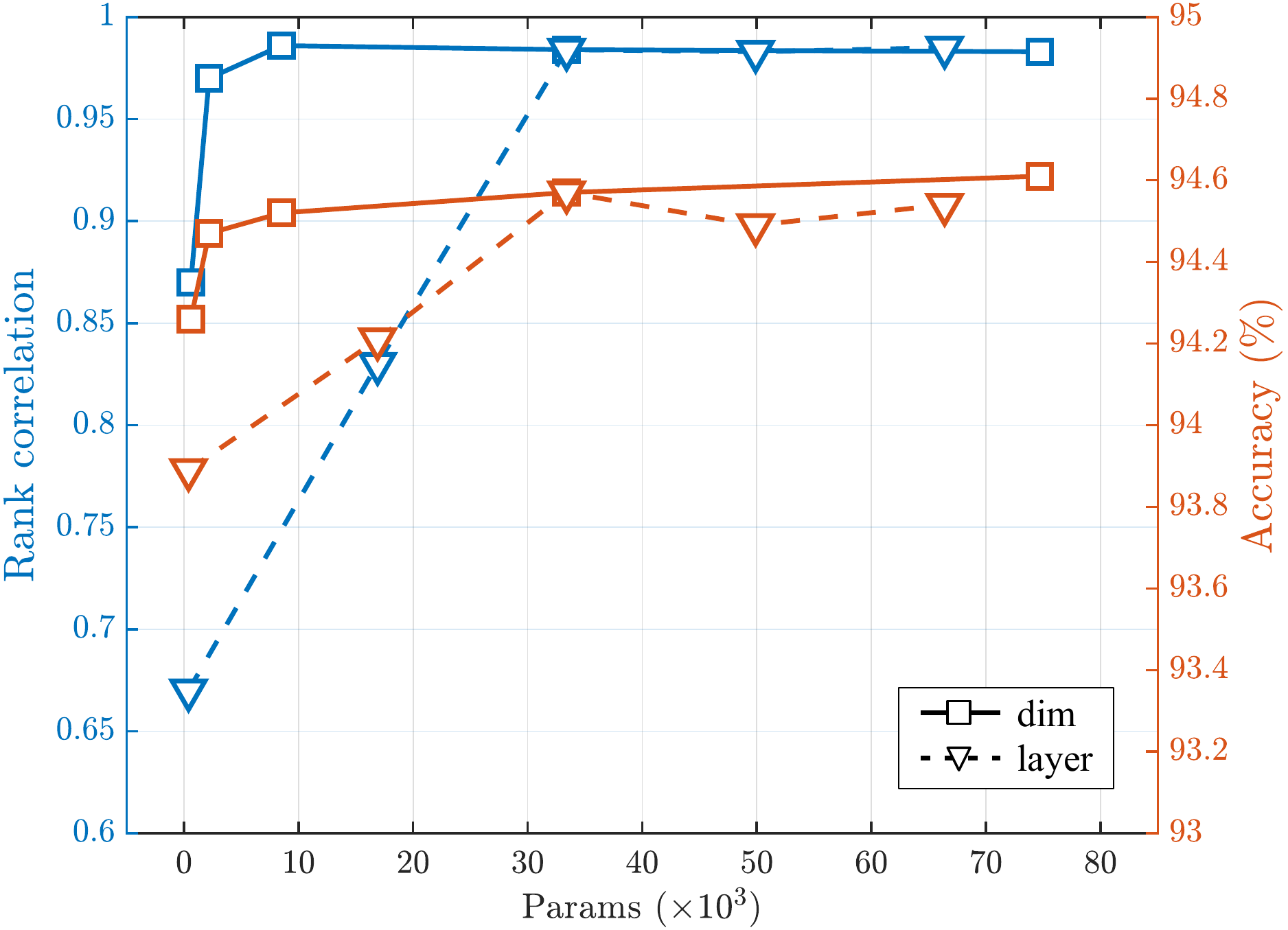}}
	\vspace{-2mm}
	\caption{(a) Evaluation results on CIFAR-10 using ReLoss with different rank correlations. (b) Evaluation results on CIFAR-10 using different capacities of surrogate losses.}
	\label{fig:ab_performance}
	\vspace{-4mm}
\end{figure}

\textbf{Performance of losses with different rank correlations.}
To show the influence of the rank correlations on the performance, we choose the surrogate losses with different rank correlations to train models on CIFAR-10 dataset. The results in Figure~\ref{fig:ab_performance} (a) clearly show that the increase of rank correlation boosts the performance, and the losses with lower rank correlations will disturb the training of networks.

\textbf{Performance of losses with different network capacities.}
Our surrogate losses are constructed with fully connected layers. To validate the influence of capacities of loss model, we conduct experiments to learn losses on different capacities on CIFAR-10 dataset. As shown in Figure~\ref{fig:ab_performance} (b), our original loss model has 33.4K parameters, we adjust the number of layers or hidden dimensions to change the network capacity. The results show that the performance of our ReLoss gets saturated on a small number of parameters. With this small network capacity, its computational cost is negligible in the training of prediction networks.

\input{tables/table_ab_alternate.tex}

\textbf{Compare with alternately learning losses.}\ \ 
The outputs of prediction networks change during the training. For surrogate losses using approximation-based optimization, it is hard to obtain accurate predictions for all the possible predictions, so the previous works~\citep{grabocka2019learning, patel2020learning} learn surrogate losses alternately with prediction networks during training. In our paper, we show that our ReLoss achieves higher performance using only pre-trained surrogate losses. We further conduct experiments to train our ReLoss alternately with prediction networks. As shown in Table~\ref{tab:ab_alt}, the alternate training obtains higher accuracies but gets more unstable on CIFAR datasets. We think this might be because the update of weights in surrogate losses will disturb the gradients, while our ReLoss without alternate training provides the same gradients for the same predictions and labels, thus more stable. Since the performance improvements of alternate training are marginal, we can use pre-trained losses for better generalization and efficiency.

\textbf{Effect of gradient penalty.}\ \ 
Our paper aims to stabilize the gradients of surrogate losses by introducing a gradient penalty regularization in loss learning. We conduct experiments to show the effectiveness of the gradient penalty. As summarized in Table~\ref{tab:ab_penalty}, we train the models on CIFAR datasets using the surrogate losses with or without gradient penalty. The results show that the ReLoss without gradient penalty performs poorly compared to the one with gradient penalty and even the original loss, although it obtains a good rank correlation. It indicates that the regularization of gradients of the surrogate losses is necessary and contributes a lot to the performance.

\vspace{-4mm}
\input{tables/table_ab_penalty.tex}

As for the weight $\lambda$ in Eq.(\ref{eq:obj_correlation_with_penalty}), we empirically find that this regularization is easy to achieve since our learning objective of correlation is weak. We have tried different values of $\lambda$, the term of gradient penalty is always very small ($\sim$1e-3), so we directly follow previous work~\citep{gulrajani2017improved} and use $\lambda = 10$.

\textbf{Robustness of performance in multiple independent runs.}\ \ 
To validate the robustness of the training of ReLoss, we conduct experiments to train the ReLoss multiple times independently, and leverage these learned surrogate losses to train ResNet-56 on CIFAR datasets. As the results summarized in Table~\ref{tab:ab_multiple_runs}, the accuracies of multiple runs are similar (with low standard variance), showing that our ReLoss can obtain stable results. We believe that our ReLoss is easy to learn, and the regularization term of gradient penalty could obtain stable gradients of surrogate losses \wrt the logits. As a result, the performance would be robust.
\input{tables/table_ab_multiple_runs.tex}

\textbf{Compare with rank-based classification loss.} Our method adopts differentiable sort algorithms~\citep{blondel2020fast,petersen2021differentiable} to train the surrogate loss. However, \citealp{blondel2020fast} proposes a rank-based classification loss to directly calculate the L1 error between predicted soft ranks and target ranks of top-1 elements, i.e.,
\begin{equation}
    l_\mathrm{rk} = |\bm{r}_\mathrm{pos}-N| ,
\end{equation}
where $\bm{r}_\mathrm{pos}$ denotes the predicted soft ranks of $\bm{y}_\mathrm{pos}$ and $N$ is the number of classes.

We train the above rank-based classification loss (RankLoss) on CIFAR-10, CIFAR-100, and ImageNet datasets with the same models and strategies in our paper. From the results summarized in Table~\ref{tab:ab_rankloss}, we can see that the accuracies obtained by RankLoss are significantly lower than the original CE loss and our ReLoss. Besides, on datasets with more classes (100 and 1000 on CIFAR-100 and ImageNet, respectively), it only obtains slightly better accuracies than random guess. One possible reason is that RankLoss only focuses on the ranks of positive elements of logits, lacking supervision on the remained elements. As a result, the network receives little information to converge on datasets with large numbers of classes. In contrast, our ReLoss learns from the evaluation metrics and supplies better information for discriminating models, thus achieves better accuracy.

On the other hand, RankLoss is hard to generalize to different tasks since it needs to design different loss functions for different metrics. Meanwhile, it cannot be applied to regression tasks.
As a result, we believe it is necessary to learn a metric-oriented surrogate loss using a neural network rather than directly applying differentiable ranking operators as the loss.

\input{tables/table_ab_rankloss.tex}

\textbf{GPU memory and training cost compared to original loss.} We report the memory consumption and training speed of ResNet-56 on CIFAR-10 and CIFAR-100 datasets in Table~\ref{tab:ab_training_cost}. Since our ReLoss only has $\sim0.03$M parameters, the memory and training time increments are negligible compared to the much larger consumptions of models (e.g., ResNet-50 has 25.6M parameters).

\input{tables/table_ab_training_cost.tex}

\subsection{Visualization of Convergence Curves in Training}
\begin{figure}[h]
	\centering
	\subfigure[CIFAR-10] 
	{\includegraphics[width=0.32\columnwidth]{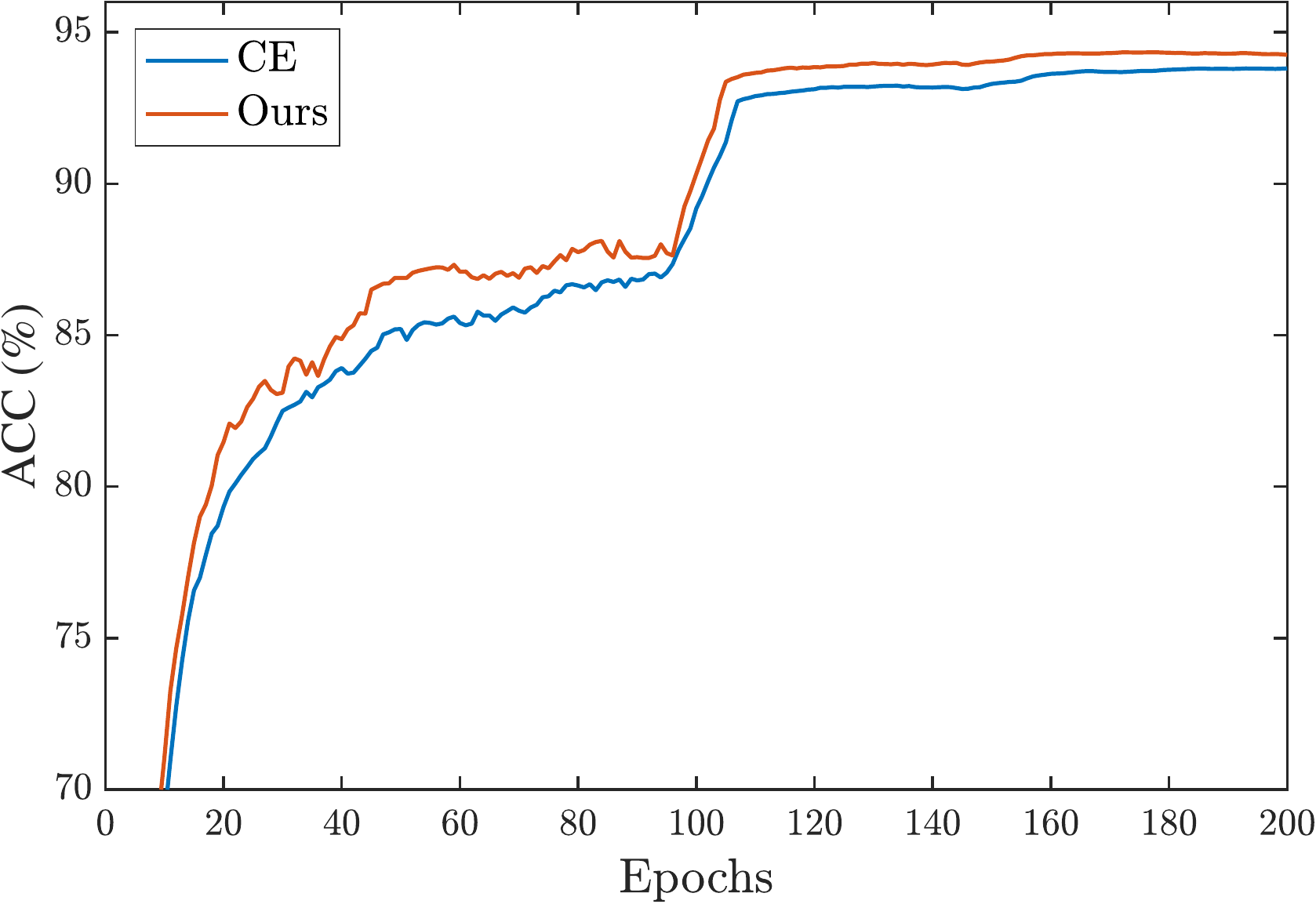}} 
	\hspace{-1mm}
	\subfigure[CIFAR-100]
	{\includegraphics[width=0.32\columnwidth]{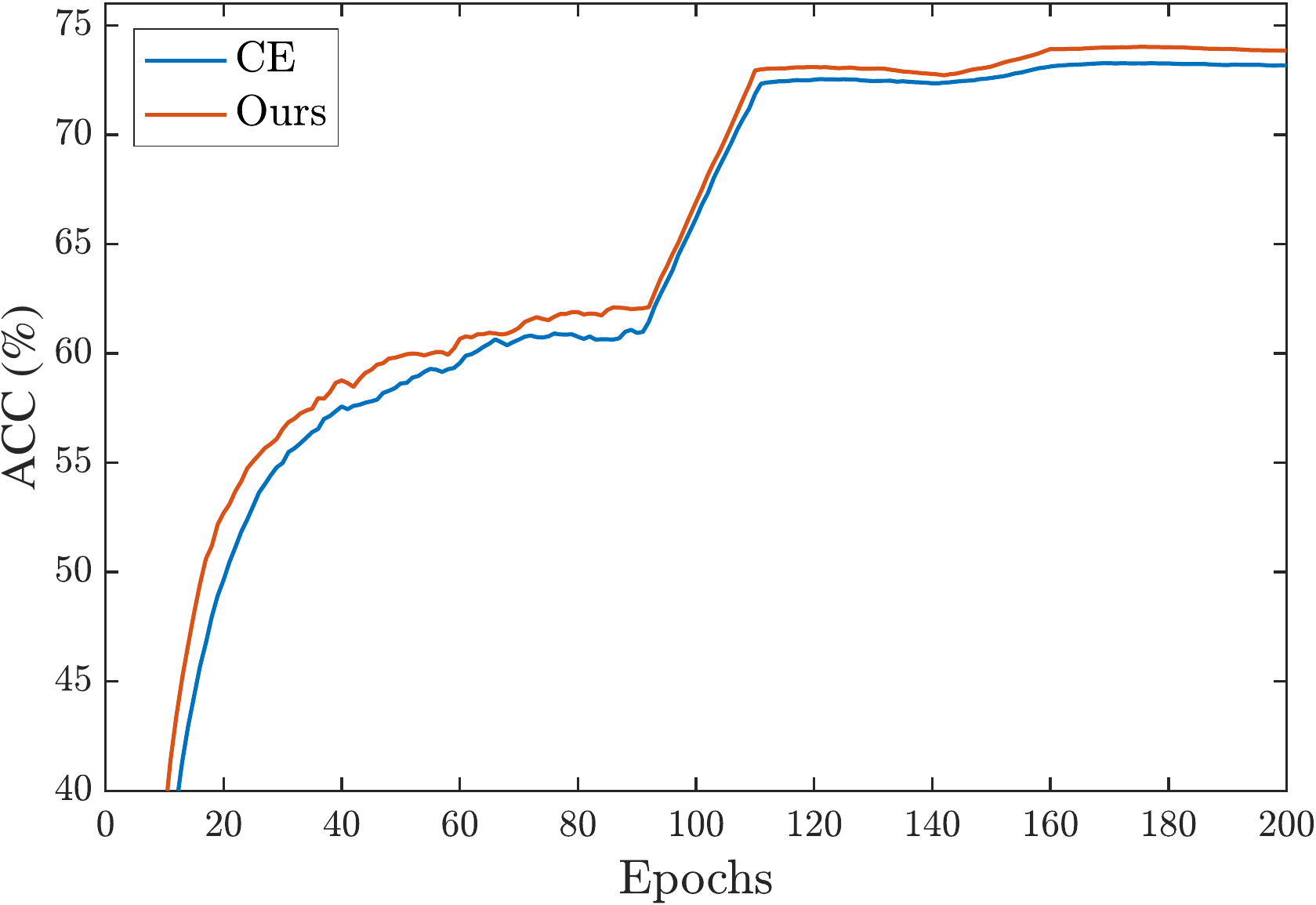}}
	\hspace{-1mm}
	\subfigure[ImageNet]
	{\includegraphics[width=0.32\columnwidth]{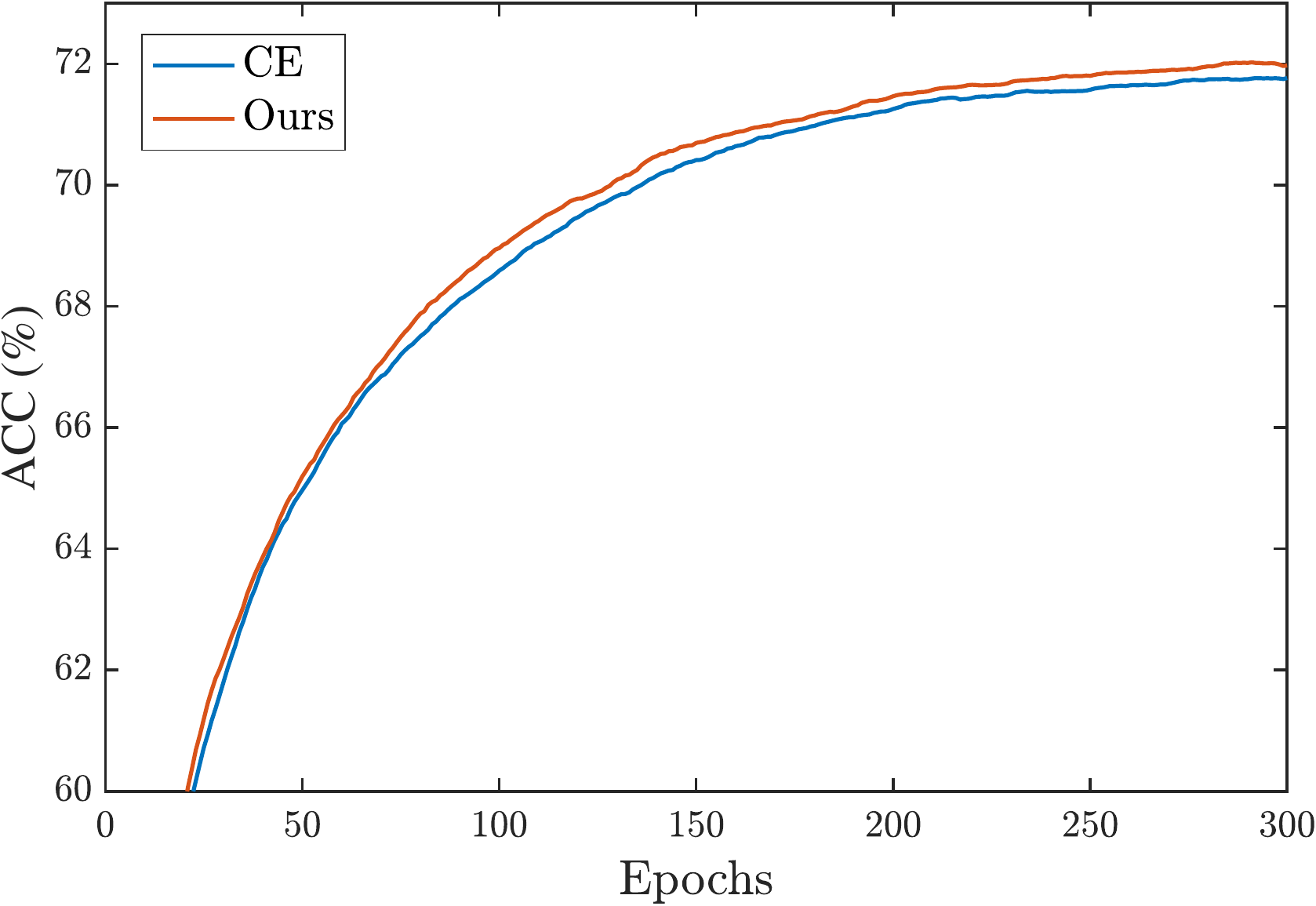}}
	\vspace{-2mm}
	\caption{Convergence curves (validation accuracies) of CE loss and our loss on CIFAR-10, CIFAR-100, and ImageNet datasets. The data is smoothed using a moving average with a factor $0.25$. Zoom up to view better.}
	\label{fig:cls_curves}
\end{figure}

We visualize the convergence curves of CE loss and our surrogate loss in Figure~\ref{fig:cls_curves}. We can see that our loss obtains higher validation accuracies over the whole training procedure.

\end{document}

%% file: math_commands.tex
\usepackage{amsmath,amsfonts,bm}

\def\eqref#1{equation~\ref{#1}}

\def\1{\bm{1}}

\DeclareMathAlphabet{\mathsfit}{\encodingdefault}{\sfdefault}{m}{sl}
\SetMathAlphabet{\mathsfit}{bold}{\encodingdefault}{\sfdefault}{bx}{n}

\DeclareMathOperator*{\argmin}{arg\,min}

%% file: alg_train_loss.tex
\begin{algorithm}[h]
	\caption{Learning of surrogate losses.}
	\label{alg:train_loss}
	
	\begin{algorithmic}[1]
	    \REQUIRE{surrogate loss $\L$ with random weights $\bm{\theta}_l$, batch size $N$, metric function $M$, data generators $G_M$ and $G_R$, sample probability $p$.}
	    \ENSURE{learned surrogate loss with highest correlation.}
		\WHILE{\textit{not converged}}
		    \STATE $L = \emptyset$ ; $M = \emptyset$ ; $L_p = \emptyset$ ;
		    \FOR{$i=1,..,N$}
		        \STATE generate a batch of predictions and ground-truth labels ($\y_i$, $\haty_i$) from $G_R$ with probability $p$ or $G_M$ with probability $1-p$ ;
		        \STATE compute loss $\wrt$ predictions and labels: $l_i = \L(\y_i, \haty_i; \bm{\theta}_l)$ ;
		        \STATE compute metric: $m_i = \M(\y_i, \haty_i)$ ;
                \STATE compute $l_{pi} = (\parallel \nabla_y \L(\y_i, \haty_i; \bm{\theta}_l) \parallel_2 - 1)^2$ ;
                \STATE $L = L \cup \{l_i\}$ ; $M = M \cup \{m_i\}$ ; $L_p = L_p \cup \{l_{pi}\}$ ;
		    \ENDFOR
		    \STATE $\L_{\textrm{penalty}} = \frac{1}{N} \sum_{i=1}^N L_p$ ;
		    \STATE optimize $\bm{\theta}_l$ by descending $\nabla_{\bm{\theta}_l}(\rho_S(L, M) + \lambda \L_{\textrm{penalty}})$ ;

		\ENDWHILE
		\RETURN learned surrogate loss with weights $\bm{\theta}_l^*$ .
	\end{algorithmic}
\end{algorithm}

%% file: tables/table_correlation.tex
\begin{table}
	\renewcommand\arraystretch{1.17}
	\setlength\tabcolsep{1.5mm}
	\centering
	\caption{Rank correlations between loss function (descending order) and metrics (ascending order) in different tasks, higher is better.}
	\label{tab:correlation}
	\footnotesize
	\begin{tabular}{l|c|c|c|c|c|c}
	    \Xhline{2\arrayrulewidth}
	    \multirow{2}*{Task} & \multirow{2}*{Metric} & Original & \multicolumn{2}{c|}{Spearman's (\%)} & \multicolumn{2}{c}{Kendall's Tau (\%)}\\
        \cline{4-7}
        ~ & ~ & loss & origin & ReLoss (ours) & origin & ReLoss (ours) \\
		\hline
		Classification & ACC & CE & 95.66 & 98.40 (\green{+2.74}) & 83.69 & 89.88 (\green{+6.19})\\
		Human Pose Estimation & PCK & MSE & 46.71 & 86.04 (\green{+39.33}) & 33.04 & 69.00 (\green{+35.96}) \\
		Machine Reading Comprehension & F1 & CE & 78.68 & 84.63 (\green{+5.95}) & 61.49 & 67.96 (\green{+6.47})\\
		Neural Machine Translation & BLEU & CE & 70.14 & 75.68 (\green{+5.54}) & 65.37 & 70.17 (\green{+4.80}) \\
		\Xhline{2\arrayrulewidth}
	\end{tabular}
	\vspace{-6mm}
\end{table}

%% file: tables/table_image_cls.tex
\begin{table}
	\renewcommand\arraystretch{1.17}
	\setlength\tabcolsep{3mm}
	\centering
	\caption{Results on CIFAR-10, CIFAR-100, and ImageNet datasets.}
	\label{tab:image_cls}
	\footnotesize
	\begin{tabular}{l|l|c|c|c|c}
		\Xhline{2\arrayrulewidth}
		\multirow{2}*{Dataset} & \multirow{2}*{Model} & \multicolumn{2}{c|}{CE} & \multicolumn{2}{c}{ReLoss} \\
		\cline{3-6}
		~ & ~ & Top-1 (\%) & Top-5 (\%) & Top-1 (\%) & Top-5 (\%) \\
		\hline
		CIFAR-10 & ResNet-56 & 94.32 $\pm$ 0.25 & - & \textbf{94.57} $\pm$ 0.08 & -\\
		CIFAR-100 & ResNet-56 & 73.61 $\pm$ 0.11 & - & \textbf{74.15} $\pm$ 0.14 & -\\
		\hline
		\multirow{2}*{ImageNet} & ResNet-50 & 76.5 & 93.0 & \textbf{76.8} & 93.0 \\
		~ & MobileNet V2 & 71.8 & 90.3 & \textbf{72.2} & 90.5\\
		\Xhline{2\arrayrulewidth}
	\end{tabular}
	\vspace{-6mm}
\end{table}

%% file: tables/table_pose_sota.tex
\begin{table}
	\renewcommand\arraystretch{1.3}
	\setlength\tabcolsep{0.6mm}
	\centering
	\caption{Results of human pose estimation task on COCO dataset.}
	\label{tab:pose}
	\small
	\begin{tabular}{l|l|c|cccccc|c}
		\Xhline{2\arrayrulewidth}
		Method & Backbone & Input size & AP & AP$^{50}$ & AP$^{75}$ & AP$^M$ & AP$^L$ & AR & PCK@0.05\\
		\hline
		\multicolumn{10}{c}{validation set}\\
		\hline
		SimpleBaseline~\citep{xiao2018simple} & ResNet-50 & $256 \times 192$ & 70.4 & 88.6 & 78.3 & 67.1 & 77.2 & 76.3 & 85.0 \\
		SimpleBaseline + ReLoss & ResNet-50 & $256 \times 192$ & \textbf{71.9} & \textbf{89.9} & \textbf{80.0} & \textbf{68.0} & \textbf{77.9} & \textbf{77.3} & \textbf{86.1} \\
		\hline
		HRNet~\citep{sun2019deep} & HRNet-W32 & $256 \times 192$ & 74.4 & \textbf{90.5} & 81.9 & 70.8 & 81.0 & 79.8 & 86.7 \\
		HRNet + ReLoss & HRNet-W32 & $256 \times 192$ & \textbf{74.8} & \textbf{90.5} & \textbf{82.4} & \textbf{70.9} & \textbf{81.2} & \textbf{79.9} & \textbf{87.3} \\
		\hline
		\multicolumn{10}{c}{test-dev set}\\
		\hline
		G-RMI~\citep{papandreou2017towards} & ResNet-101 & $353 \times 257$ & 64.9 & 85.5 & 71.3 & 62.3 & 70.0 & 69.7 & - \\
		SimpleBaseline~\citep{xiao2018simple} & ResNet-101 & $384 \times 288$ & 73.7 & 91.9 & 81.1 & 70.3 & 80.0 & 79.0 & - \\
		HRNet~\citep{sun2019deep} & HRNet-W48 & $384 \times 288$ & 75.5 & 92.5 & 83.3 & 71.9 & 81.5 & 80.5 & - \\
		DARK~\citep{zhang2020distribution} & HRNet-W48 & $384 \times 288$ & 76.2 & 92.5 & 83.6 & 72.5 & 82.4 & 81.1 & -\\
		DARK + ReLoss & HRNet-W48 & $384 \times 288$ & \textbf{76.4} & \textbf{92.7} & \textbf{83.7} & \textbf{72.7} & \textbf{82.5} & \textbf{81.3} & -\\
		\Xhline{2\arrayrulewidth}
	\end{tabular}
	\vspace{-4mm}
\end{table}

%% file: tables/table_mrc_sota.tex
\begin{minipage}{\textwidth}
    \begin{minipage}[c]{0.60\textwidth}
    \centering
    	\renewcommand\arraystretch{1.2}
    	\setlength\tabcolsep{1mm}
    	\centering
    	\makeatletter\def\@captype{table}\makeatother\caption{Results of machine reading comprehension task on DuReader 2.0 dataset. $\dag$: reported by \citep{he2018dureader}.}
    	\label{tab:mrc}
    	\footnotesize
    	\begin{tabular}{l|c|c|c}
    		\Xhline{2\arrayrulewidth}
    		Method & ROUGE-L & BLEU-4 & F1\\
    		\hline
    		\multicolumn{4}{c}{dev set}\\
    		\hline
    		MacBERT-base~\citep{cui2020revisiting} & 51.4 & 50.3 & 53.9\\
    		MacBERT-base + ReLoss & \textbf{51.8} & \textbf{50.6} & \textbf{54.2}\\
    		\hline
    		MacBERT-large~\citep{cui2020revisiting} & 53.2 & 51.2 & 55.5 \\
    		MacBERT-large + ReLoss & \textbf{53.6} & \textbf{51.4} & \textbf{55.9} \\
    		\hline
    		\multicolumn{4}{c}{test set}\\
    		\hline
    		BiDAF$^\dag$~\citep{seo2016bidirectional} & 39.2 & 31.9 & - \\
    		\cite{wang2018multi} & 44.2 & 41.0 & -\\
    		MCR-Net-large~\citep{peng2021mcr} & 50.8 & 49.2 & -\\
    		\hline
    		Human Performance$^\dag$ & 57.4 & 56.1 & - \\
    		\hline
    		MacBERT-large + ReLoss & \textbf{64.9} & \textbf{61.8} & - \\
    		\Xhline{2\arrayrulewidth}
    	\end{tabular}
    \end{minipage}
    \hspace{6mm}
    \begin{minipage}[c]{0.30\textwidth}
    	\renewcommand\arraystretch{1.5}
    	\setlength\tabcolsep{1mm}
    	\centering
    	\makeatletter\def\@captype{table}\makeatother\caption{Results on SQuAD 1.1 dataset compared with BERT~\citep{devlin2018bert}. }
    	\label{tab:mrc_squad}
    	\footnotesize
    	\begin{tabular}{l|c|c}
    		\Xhline{2\arrayrulewidth}
    		Method & F1 & EM\\
    		\hline
    		BERT-base & 88.5 & 80.8\\
    		BERT-base + ReLoss & \textbf{88.8} & \textbf{81.3} \\
    		\hline
    		BERT-large & 90.9 & 84.1 \\
    		BERT-large + ReLoss & \textbf{91.4} & \textbf{84.6} \\
    		\Xhline{2\arrayrulewidth}
    	\end{tabular}
    \end{minipage}
\end{minipage}

%% file: tables/table_nat.tex
\begin{table}
	\renewcommand\arraystretch{1.13}
	\setlength\tabcolsep{1mm}
	\centering
	\caption{Evaluation results of BLEU on Neural Machine Translation task. We report the performance of our methods on the WMT16 EN-RO dataset. Transformer denotes the auto-regressive model. * denotes the performance that we reproduced using the public code. }
	\label{tab:nat}
	\footnotesize
	\begin{tabular}{l|c|cc|cc|cc}
		\Xhline{2\arrayrulewidth}
		\multirow{2}*{Model} & \multirow{2}*{Speed} & \multicolumn{2}{c|}{Original loss} & \multicolumn{2}{c|}{ReLoss on EN-RO} & \multicolumn{2}{c}{ReLoss on RO-EN} \\
		\cline{3-8}
        ~ & ~ & EN-RO & RO-EN & EN-RO & RO-EN & EN-RO & RO-EN\\
		\hline
		Transformer~\citep{vaswani2017attention} & 1.0$\times$ & 32.88 & 33.94 & - & - & - & -\\
		\hline
		NAT-Base~\citep{DBLP:journals/corr/abs-1711-02281} & 15.6$\times$ & 29.24 & 28.97 & 30.07 \tiny\green{+0.83} & 29.68 \tiny\green{+0.71} & 29.93 \tiny\green{+0.69} & 29.61 \tiny\green{+0.64}\\
		BoN-$L_{1}$(N=2)$^*$~\citep{DBLP:journals/corr/abs-2106-08122} & 15.6$\times$ & 30.76 & 30.46 & 30.96 \tiny\green{+0.20} & 30.74 \tiny\green{+0.28} & 30.88 \tiny\green{+0.12} & 30.78 \tiny\green{+0.32} \\
		\Xhline{2\arrayrulewidth}
	\end{tabular}
	\vspace{-4mm}
\end{table}

%% file: tables/table_str.tex
\begin{table}[htbp]
	\renewcommand\arraystretch{1.1}
	\setlength\tabcolsep{1.0mm}
	\centering
	\caption{Evaluation results on scene text recognition task comparing with CE and LS-ED. The reported metrics are accuracy (ACC, higher is better), normalized edit distance (NED, higher is better), and total edit distance (TED, lower is better).}
	\label{tab:str}
	\footnotesize
	\begin{tabular}{l|ccc|ccc|ccc}
		\Xhline{2\arrayrulewidth}
		\multirow{2}*{Test dataset} & \multicolumn{3}{c|}{$\uparrow$ACC (\%)} & \multicolumn{3}{c|}{$\uparrow$NED} & \multicolumn{3}{c}{$\downarrow$TED} \\
		\cline{2-10}
		~ & CE & LS-ED & ReLoss & CE & LS-ED & ReLoss & CE & LS-ED & ReLoss \\
		\hline
		IIIT-5K~\citep{mishra2012scene} & 87.500 & \textbf{87.933} & 87.700 & 0.961 & \textbf{0.963} & 0.961 & 722 & \textbf{645} & 667\\
		SVT~\citep{wang2011end} & 87.172 & 86.708 & \textbf{87.481} & 0.952 & 0.954 & \textbf{0.957} & 180 & 163 & \textbf{156}\\
		ICDAR'03~\citep{lucas2005icdar} & 94.302 & 94.535 & \textbf{94.579} & 0.979 & 0.981 & \textbf{0.982} & 110 & 99 & \textbf{98}\\
		ICDAR'13~\citep{karatzas2013icdar} & 92.020 & 92.299 & \textbf{92.709} & 0.966 & 0.979 & \textbf{0.981} & 137 & 108 & \textbf{101}\\
		ICDAR'15~\citep{karatzas2015icdar} & \textbf{78.520} & 78.410 & 78.355 & 0.915 & 0.915 & \textbf{0.916} & 868 & \textbf{837} & 845\\
		SVTP~\citep{phan2013recognizing} & 78.605 & 79.225 & \textbf{80.310} & 0.912 & 0.913 & \textbf{0.915} & 346 & 333 & \textbf{316}\\
		CUTE~\citep{risnumawan2014robust} & 73.171 & 74.216 & \textbf{75.958} & 0.871 & 0.875 & \textbf{0.884} & 224 & 219 & \textbf{195}\\
		\hline
		Wins & 1 & 1 & \textbf{5} & 0 & 1 & \textbf{6} & 0 & 2 & \textbf{5} \\
		Average & 84.470 & 84.761 & \textbf{85.299} & 0.937 & 0.940 & \textbf{0.943} & 370 & 343 & \textbf{340} \\
		\Xhline{2\arrayrulewidth}
	\end{tabular}
\end{table}

%% file: tables/table_ab_opt.tex
\begin{table}[htbp]
	\renewcommand\arraystretch{1.17}
	\setlength\tabcolsep{2mm}
	\centering
	\vspace{-4mm}
	\caption{Results of different optimization methods on image classification task.}
	\label{tab:ab_opt}
	\footnotesize
	\begin{tabular}{l|c|c|c|c}
		\Xhline{2\arrayrulewidth}
        \multirow{2}*{Loss function} & \multicolumn{2}{c|}{Rank correlation (\%)} & \multicolumn{2}{c}{ACC (\%)}  \\
		\cline{2-5}
		~ & Spearman's & Kendall's Tau & CIFAR-10 & CIFAR-100 \\
		\hline
		Cross Entropy & 95.66 & 83.69 & 94.32 $\pm$ 0.25 & 73.61 $\pm$ 0.11 \\ 
		ReLoss (approximation-based) & 91.71 & 76.03 & 94.11 $\pm$ 0.42 & 73.88 $\pm$ 0.32 \\ 
		ReLoss (correlation-based) & \textbf{98.40} & \textbf{89.88} &  \textbf{94.57} $\pm$ 0.08 & \textbf{74.15} $\pm$ 0.14 \\
		\Xhline{2\arrayrulewidth}
	\end{tabular}
	\vspace{-2mm}
\end{table}

%% file: tables/table_ab_alternate.tex
\begin{table}[h]
	\renewcommand\arraystretch{1.17}
	\setlength\tabcolsep{2mm}
	\centering
	\caption{Compare with alternate training on image classification.}
	\label{tab:ab_alt}
	\footnotesize
	\begin{tabular}{l|c|c|c}
		\Xhline{2\arrayrulewidth}
         \multirow{2}*{Loss function} & \multicolumn{3}{c}{ACC (\%)}  \\
		\cline{2-4}
		~ & CIFAR-10 & CIFAR-100 & ImageNet \\
		\hline
		Cross Entropy & 94.32 $\pm$ 0.25 & 73.61 $\pm$ 0.11 & 76.4 \\
		ReLoss & 94.57 $\pm$ 0.08 & 74.15 $\pm$ 0.14 & 76.8 \\
		ReLoss (alternate training) & \textbf{94.65} $\pm$ 0.21 & \textbf{74.18} $\pm$ 0.31 & \textbf{76.9} \\
		\Xhline{2\arrayrulewidth}
	\end{tabular}
	\vspace{0mm}
\end{table}

%% file: tables/table_ab_penalty.tex
\begin{table}[h]
	\renewcommand\arraystretch{1.17}
	\setlength\tabcolsep{2mm}
	\centering
	\caption{Results of ReLoss with or without gradient penalty.}
	\label{tab:ab_penalty}
	\footnotesize
	\begin{tabular}{l|c|c|c|c}
		\Xhline{2\arrayrulewidth}
         \multirow{2}*{Loss function} & \multicolumn{2}{c|}{Rank correlation (\%)} & \multicolumn{2}{c}{ACC (\%)}  \\
		\cline{2-5}
		~ & Spearman's & Kendall's Tau & CIFAR-10 & CIFAR-100 \\
		\hline
		Cross Entropy & 95.66 & 83.69 & 94.32 $\pm$ 0.25 & 73.61 $\pm$ 0.11 \\
		ReLoss (w/o gradient penalty) & 98.31 & 89.56 & 94.28 $\pm$ 0.31 & 73.03 $\pm$ 0.26\\
		ReLoss (w/ gradient penalty) & \textbf{98.40} & \textbf{89.88} &  \textbf{94.57} $\pm$ 0.08 & \textbf{74.15} $\pm$ 0.14 \\
		\Xhline{2\arrayrulewidth}
	\end{tabular}
\end{table}

%% file: tables/table_ab_multiple_runs.tex
\begin{table}[h]
	\renewcommand\arraystretch{1.17}
	\setlength\tabcolsep{4mm}
	\centering
	\vspace{-4mm}
	\caption{Results of ReLoss on CIFAR datasets in multiple independent runs.}
	\label{tab:ab_multiple_runs}
	\footnotesize
	\begin{tabular}{c|c|c|c|c}
		\Xhline{2\arrayrulewidth}
		\multirow{2}*{Number} & \multicolumn{2}{c|}{Rank correlation (\%)} & \multicolumn{2}{c}{ACC (\%)}  \\
		\cline{2-5}
		~ & Spearman's & Kendall's Tau & CIFAR-10 & CIFAR-100 \\
		\hline
		1 & 98.36 & 89.91 & 94.49 $\pm$ 0.06 & 74.05 $\pm$ 0.09 \\
		2 & 98.47 & 89.78 & 94.51 $\pm$ 0.08 & 74.12 $\pm$ 0.11 \\
		3 & 98.43 & 89.86 & 94.43 $\pm$ 0.10 & 74.09 $\pm$ 0.08 \\
		4 & 98.42 & 89.75 & 94.59 $\pm$ 0.08 & 74.15 $\pm$ 0.10 \\
		5 & 98.38 & 89.71 & 94.55 $\pm$ 0.07 & 74.11 $\pm$ 0.06 \\
		\hline
		mean $\pm$ std & 98.41 $\pm$ 0.04 & 89.80 $\pm$ 0.08 & 94.51 $\pm$ 0.06 & 74.10 $\pm$ 0.03\\
		\Xhline{2\arrayrulewidth}
	\end{tabular}
	\vspace{-2mm}
\end{table}

%% file: tables/table_ab_rankloss.tex
\begin{table}[h]
	\renewcommand\arraystretch{1.2}
	\setlength\tabcolsep{3mm}
	\centering
	\caption{Compare with rank-based classification loss.}
	\label{tab:ab_rankloss}
	\footnotesize
	\begin{tabular}{l|c|c|c|c}
		\Xhline{2\arrayrulewidth}
        Dataset & Model & CE Loss (\%) & RankLoss (\%) & ReLoss (\%)\\
		\hline
		CIFAR-10 & ResNet-56 & 94.32 & 82.77 & \textbf{94.57}\\
		CIFAR-100 & ResNet-56 & 73.61 & 5.65 & \textbf{74.15}\\
		ImageNet & ResNet-50 & 76.5 & 0.58 & \textbf{76.8}\\
		\Xhline{2\arrayrulewidth}
	\end{tabular}
\end{table}

%% file: tables/table_ab_training_cost.tex
\begin{table}[h]
	\renewcommand\arraystretch{1.17}
	\setlength\tabcolsep{3mm}
	\centering
	\vspace{-4mm}
	\caption{Comparisons of GPU memory and training cost.}
	\label{tab:ab_training_cost}
	\footnotesize
	\begin{tabular}{c|c|c|c|c}
		\Xhline{2\arrayrulewidth}
		Dataset & Loss function & GPU memory (M) & \thead{Training speed\\(batches / second)} & ACC (\%)\\
		\hline
		\multirow{2}*{CIFAR-10} & CE & 618.46 & 14.29 & 94.32\\
		~ & ReLoss & 618.68 & 14.29 & \textbf{94.57}\\
		\hline
		\multirow{2}*{CIFAR-100} & CE & 618.55 & 14.27 & 73.61\\
		~ & ReLoss & 618.76 & 14.27 & \textbf{74.15}\\
		\Xhline{2\arrayrulewidth}
	\end{tabular}
	\vspace{-2mm}
\end{table}

%% file: main.bbl
\begin{thebibliography}{47}
\providecommand{\natexlab}[1]{#1}
\providecommand{\url}[1]{\texttt{#1}}
\expandafter\ifx\csname urlstyle\endcsname\relax
  \providecommand{\doi}[1]{doi: #1}\else
  \providecommand{\doi}{doi: \begingroup \urlstyle{rm}\Url}\fi

\bibitem[Adams \& Zemel(2011)Adams and Zemel]{adams2011ranking}
Ryan~Prescott Adams and Richard~S Zemel.
\newblock Ranking via sinkhorn propagation.
\newblock \emph{arXiv preprint arXiv:1106.1925}, 2011.

\bibitem[Blondel et~al.(2020)Blondel, Teboul, Berthet, and
  Djolonga]{blondel2020fast}
Mathieu Blondel, Olivier Teboul, Quentin Berthet, and Josip Djolonga.
\newblock Fast differentiable sorting and ranking.
\newblock In \emph{International Conference on Machine Learning}, pp.\
  950--959. PMLR, 2020.

\bibitem[Clevert et~al.(2015)Clevert, Unterthiner, and
  Hochreiter]{clevert2015fast}
Djork-Arn{\'e} Clevert, Thomas Unterthiner, and Sepp Hochreiter.
\newblock Fast and accurate deep network learning by exponential linear units
  (elus).
\newblock \emph{arXiv preprint arXiv:1511.07289}, 2015.

\bibitem[Contributors(2020)]{mmpose2020}
MMPose Contributors.
\newblock Openmmlab pose estimation toolbox and benchmark.
\newblock \url{https://github.com/open-mmlab/mmpose}, 2020.

\bibitem[Cui et~al.(2020)Cui, Che, Liu, Qin, Wang, and Hu]{cui2020revisiting}
Yiming Cui, Wanxiang Che, Ting Liu, Bing Qin, Shijin Wang, and Guoping Hu.
\newblock Revisiting pre-trained models for chinese natural language
  processing.
\newblock \emph{arXiv preprint arXiv:2004.13922}, 2020.

\bibitem[Deng et~al.(2009)Deng, Dong, Socher, Li, Li, and Fei-Fei]{Imagenet}
Jia Deng, Wei Dong, Richard Socher, Li-Jia Li, Kai Li, and Li~Fei-Fei.
\newblock Imagenet: A large-scale hierarchical image database.
\newblock In \emph{2009 IEEE conference on computer vision and pattern
  recognition}, pp.\  248--255. Ieee, 2009.

\bibitem[Devlin et~al.(2018)Devlin, Chang, Lee, and Toutanova]{devlin2018bert}
Jacob Devlin, Ming-Wei Chang, Kenton Lee, and Kristina Toutanova.
\newblock Bert: Pre-training of deep bidirectional transformers for language
  understanding.
\newblock \emph{arXiv preprint arXiv:1810.04805}, 2018.

\bibitem[DeVries \& Taylor(2017)DeVries and Taylor]{devries2017improved}
Terrance DeVries and Graham~W Taylor.
\newblock Improved regularization of convolutional neural networks with cutout.
\newblock \emph{arXiv preprint arXiv:1708.04552}, 2017.

\bibitem[Dodge(2008)]{dodge2008concise}
Yadolah Dodge.
\newblock \emph{The concise encyclopedia of statistics}.
\newblock Springer Science \& Business Media, 2008.

\bibitem[Grabocka et~al.(2019)Grabocka, Scholz, and
  Schmidt-Thieme]{grabocka2019learning}
Josif Grabocka, Randolf Scholz, and Lars Schmidt-Thieme.
\newblock Learning surrogate losses.
\newblock \emph{arXiv preprint arXiv:1905.10108}, 2019.

\bibitem[Grover et~al.(2018)Grover, Wang, Zweig, and
  Ermon]{grover2018stochastic}
Aditya Grover, Eric Wang, Aaron Zweig, and Stefano Ermon.
\newblock Stochastic optimization of sorting networks via continuous
  relaxations.
\newblock In \emph{International Conference on Learning Representations}, 2018.

\bibitem[Gu et~al.(2017)Gu, Bradbury, Xiong, Li, and
  Socher]{DBLP:journals/corr/abs-1711-02281}
Jiatao Gu, James Bradbury, Caiming Xiong, Victor O.~K. Li, and Richard Socher.
\newblock Non-autoregressive neural machine translation.
\newblock \emph{CoRR}, abs/1711.02281, 2017.
\newblock URL \url{http://arxiv.org/abs/1711.02281}.

\bibitem[Gulrajani et~al.(2017)Gulrajani, Ahmed, Arjovsky, Dumoulin, and
  Courville]{gulrajani2017improved}
Ishaan Gulrajani, Faruk Ahmed, Martin Arjovsky, Vincent Dumoulin, and Aaron
  Courville.
\newblock Improved training of wasserstein gans.
\newblock In \emph{Proceedings of the 31st International Conference on Neural
  Information Processing Systems}, pp.\  5769--5779, 2017.

\bibitem[He et~al.(2016)He, Zhang, Ren, and Sun]{he2016deep}
Kaiming He, Xiangyu Zhang, Shaoqing Ren, and Jian Sun.
\newblock Deep residual learning for image recognition.
\newblock In \emph{Proceedings of the IEEE conference on computer vision and
  pattern recognition}, pp.\  770--778, 2016.

\bibitem[He et~al.(2018)He, Liu, Liu, Lyu, Zhao, Xiao, Liu, Wang, Wu, She,
  et~al.]{he2018dureader}
Wei He, Kai Liu, Jing Liu, Yajuan Lyu, Shiqi Zhao, Xinyan Xiao, Yuan Liu,
  Yizhong Wang, Hua Wu, Qiaoqiao She, et~al.
\newblock Dureader: a chinese machine reading comprehension dataset from
  real-world applications.
\newblock In \emph{Proceedings of the Workshop on Machine Reading for Question
  Answering}, pp.\  37--46, 2018.

\bibitem[Hinton et~al.(2015)Hinton, Vinyals, and
  Dean]{DBLP:journals/corr/HintonVD15}
Geoffrey~E. Hinton, Oriol Vinyals, and Jeffrey Dean.
\newblock Distilling the knowledge in a neural network.
\newblock \emph{CoRR}, abs/1503.02531, 2015.
\newblock URL \url{http://arxiv.org/abs/1503.02531}.

\bibitem[Karatzas et~al.(2013)Karatzas, Shafait, Uchida, Iwamura, i~Bigorda,
  Mestre, Mas, Mota, Almazan, and De~Las~Heras]{karatzas2013icdar}
Dimosthenis Karatzas, Faisal Shafait, Seiichi Uchida, Masakazu Iwamura,
  Lluis~Gomez i~Bigorda, Sergi~Robles Mestre, Joan Mas, David~Fernandez Mota,
  Jon~Almazan Almazan, and Lluis~Pere De~Las~Heras.
\newblock Icdar 2013 robust reading competition.
\newblock In \emph{2013 12th International Conference on Document Analysis and
  Recognition}, pp.\  1484--1493. IEEE, 2013.

\bibitem[Karatzas et~al.(2015)Karatzas, Gomez-Bigorda, Nicolaou, Ghosh,
  Bagdanov, Iwamura, Matas, Neumann, Chandrasekhar, Lu,
  et~al.]{karatzas2015icdar}
Dimosthenis Karatzas, Lluis Gomez-Bigorda, Anguelos Nicolaou, Suman Ghosh,
  Andrew Bagdanov, Masakazu Iwamura, Jiri Matas, Lukas Neumann,
  Vijay~Ramaseshan Chandrasekhar, Shijian Lu, et~al.
\newblock Icdar 2015 competition on robust reading.
\newblock In \emph{2015 13th International Conference on Document Analysis and
  Recognition (ICDAR)}, pp.\  1156--1160. IEEE, 2015.

\bibitem[Kendall(1938)]{kendall1938new}
Maurice~G Kendall.
\newblock A new measure of rank correlation.
\newblock \emph{Biometrika}, 30\penalty0 (1/2):\penalty0 81--93, 1938.

\bibitem[Kim \& Rush(2016)Kim and Rush]{DBLP:conf/emnlp/KimR16}
Yoon Kim and Alexander~M. Rush.
\newblock Sequence-level knowledge distillation.
\newblock In Jian Su, Xavier Carreras, and Kevin Duh (eds.), \emph{Proceedings
  of the 2016 Conference on Empirical Methods in Natural Language Processing,
  {EMNLP} 2016, Austin, Texas, USA, November 1-4, 2016}, pp.\  1317--1327. The
  Association for Computational Linguistics, 2016.
\newblock \doi{10.18653/v1/d16-1139}.
\newblock URL \url{https://doi.org/10.18653/v1/d16-1139}.

\bibitem[Krizhevsky et~al.(2009)Krizhevsky, Hinton,
  et~al.]{krizhevsky2009learning}
Alex Krizhevsky, Geoffrey Hinton, et~al.
\newblock Learning multiple layers of features from tiny images.
\newblock 2009.

\bibitem[Lin(2004)]{lin2004rouge}
Chin-Yew Lin.
\newblock Rouge: A package for automatic evaluation of summaries.
\newblock In \emph{Text summarization branches out}, pp.\  74--81, 2004.

\bibitem[Lin et~al.(2014)Lin, Maire, Belongie, Hays, Perona, Ramanan,
  Doll{\'a}r, and Zitnick]{lin2014microsoft}
Tsung-Yi Lin, Michael Maire, Serge Belongie, James Hays, Pietro Perona, Deva
  Ramanan, Piotr Doll{\'a}r, and C~Lawrence Zitnick.
\newblock Microsoft coco: Common objects in context.
\newblock In \emph{European conference on computer vision}, pp.\  740--755.
  Springer, 2014.

\bibitem[Lucas et~al.(2005)Lucas, Panaretos, Sosa, Tang, Wong, Young, Ashida,
  Nagai, Okamoto, Yamamoto, et~al.]{lucas2005icdar}
Simon~M Lucas, Alex Panaretos, Luis Sosa, Anthony Tang, Shirley Wong, Robert
  Young, Kazuki Ashida, Hiroki Nagai, Masayuki Okamoto, Hiroaki Yamamoto,
  et~al.
\newblock Icdar 2003 robust reading competitions: entries, results, and future
  directions.
\newblock \emph{International Journal of Document Analysis and Recognition
  (IJDAR)}, 7\penalty0 (2-3):\penalty0 105--122, 2005.

\bibitem[Marcel \& Rodriguez(2010)Marcel and Rodriguez]{marcel2010torchvision}
S{\'e}bastien Marcel and Yann Rodriguez.
\newblock Torchvision the machine-vision package of torch.
\newblock In \emph{Proceedings of the 18th ACM international conference on
  Multimedia}, pp.\  1485--1488, 2010.

\bibitem[Mishra et~al.(2012)Mishra, Alahari, and Jawahar]{mishra2012scene}
Anand Mishra, Karteek Alahari, and CV~Jawahar.
\newblock Scene text recognition using higher order language priors.
\newblock In \emph{BMVC-British Machine Vision Conference}. BMVA, 2012.

\bibitem[Papandreou et~al.(2017)Papandreou, Zhu, Kanazawa, Toshev, Tompson,
  Bregler, and Murphy]{papandreou2017towards}
George Papandreou, Tyler Zhu, Nori Kanazawa, Alexander Toshev, Jonathan
  Tompson, Chris Bregler, and Kevin Murphy.
\newblock Towards accurate multi-person pose estimation in the wild.
\newblock In \emph{Proceedings of the IEEE conference on computer vision and
  pattern recognition}, pp.\  4903--4911, 2017.

\bibitem[Papineni et~al.(2002)Papineni, Roukos, Ward, and
  Zhu]{DBLP:conf/acl/PapineniRWZ02}
Kishore Papineni, Salim Roukos, Todd Ward, and Wei{-}Jing Zhu.
\newblock Bleu: a method for automatic evaluation of machine translation.
\newblock In \emph{Proceedings of the 40th Annual Meeting of the Association
  for Computational Linguistics, July 6-12, 2002, Philadelphia, PA, {USA}},
  pp.\  311--318. {ACL}, 2002.
\newblock \doi{10.3115/1073083.1073135}.
\newblock URL \url{https://aclanthology.org/P02-1040/}.

\bibitem[Patel et~al.(2020)Patel, Hoda{\v{n}}, and Matas]{patel2020learning}
Yash Patel, Tom{\'a}{\v{s}} Hoda{\v{n}}, and Ji{\v{r}}{\'\i} Matas.
\newblock Learning surrogates via deep embedding.
\newblock In \emph{European Conference on Computer Vision}, pp.\  205--221.
  Springer, 2020.

\bibitem[Peng et~al.(2021)Peng, Hu, Yu, Xing, Xie, Zhu, and Sun]{peng2021mcr}
Wei Peng, Yue Hu, Jing Yu, Luxi Xing, Yuqiang Xie, Zihao Zhu, and Yajing Sun.
\newblock Mcr-net: A multi-step co-interactive relation network for
  unanswerable questions on machine reading comprehension.
\newblock In \emph{ICASSP 2021-2021 IEEE International Conference on Acoustics,
  Speech and Signal Processing (ICASSP)}, pp.\  7818--7822. IEEE, 2021.

\bibitem[Petersen et~al.(2021)Petersen, Borgelt, Kuehne, and
  Deussen]{petersen2021differentiable}
Felix Petersen, Christian Borgelt, Hilde Kuehne, and Oliver Deussen.
\newblock Differentiable sorting networks for scalable sorting and ranking
  supervision.
\newblock \emph{arXiv preprint arXiv:2105.04019}, 2021.

\bibitem[Phan et~al.(2013)Phan, Shivakumara, Tian, and
  Tan]{phan2013recognizing}
Trung~Quy Phan, Palaiahnakote Shivakumara, Shangxuan Tian, and Chew~Lim Tan.
\newblock Recognizing text with perspective distortion in natural scenes.
\newblock In \emph{Proceedings of the IEEE International Conference on Computer
  Vision}, pp.\  569--576, 2013.

\bibitem[Rajpurkar et~al.(2016)Rajpurkar, Zhang, Lopyrev, and
  Liang]{rajpurkar2016squad}
Pranav Rajpurkar, Jian Zhang, Konstantin Lopyrev, and Percy Liang.
\newblock Squad: 100,000+ questions for machine comprehension of text.
\newblock In \emph{Proceedings of the 2016 Conference on Empirical Methods in
  Natural Language Processing}, pp.\  2383--2392, 2016.

\bibitem[Risnumawan et~al.(2014)Risnumawan, Shivakumara, Chan, and
  Tan]{risnumawan2014robust}
Anhar Risnumawan, Palaiahankote Shivakumara, Chee~Seng Chan, and Chew~Lim Tan.
\newblock A robust arbitrary text detection system for natural scene images.
\newblock \emph{Expert Systems with Applications}, 41\penalty0 (18):\penalty0
  8027--8048, 2014.

\bibitem[Sandler et~al.(2018)Sandler, Howard, Zhu, Zhmoginov, and
  Chen]{sandler2018mobilenetv2}
Mark Sandler, Andrew Howard, Menglong Zhu, Andrey Zhmoginov, and Liang-Chieh
  Chen.
\newblock Mobilenetv2: Inverted residuals and linear bottlenecks.
\newblock In \emph{Proceedings of the IEEE conference on computer vision and
  pattern recognition}, pp.\  4510--4520, 2018.

\bibitem[Seo et~al.(2016)Seo, Kembhavi, Farhadi, and
  Hajishirzi]{seo2016bidirectional}
Minjoon Seo, Aniruddha Kembhavi, Ali Farhadi, and Hannaneh Hajishirzi.
\newblock Bidirectional attention flow for machine comprehension.
\newblock \emph{arXiv preprint arXiv:1611.01603}, 2016.

\bibitem[Shao et~al.(2021)Shao, Feng, Zhang, Meng, and
  Zhou]{DBLP:journals/corr/abs-2106-08122}
Chenze Shao, Yang Feng, Jinchao Zhang, Fandong Meng, and Jie Zhou.
\newblock Sequence-level training for non-autoregressive neural machine
  translation.
\newblock \emph{CoRR}, abs/2106.08122, 2021.
\newblock URL \url{https://arxiv.org/abs/2106.08122}.

\bibitem[Su et~al.(2020)Su, You, Huang, Wang, Qian, Zhang, and
  Xu]{su2020locally}
Xiu Su, Shan You, Tao Huang, Fei Wang, Chen Qian, Changshui Zhang, and Chang
  Xu.
\newblock Locally free weight sharing for network width search.
\newblock In \emph{International Conference on Learning Representations}, 2020.

\bibitem[Sun et~al.(2019)Sun, Xiao, Liu, and Wang]{sun2019deep}
Ke~Sun, Bin Xiao, Dong Liu, and Jingdong Wang.
\newblock Deep high-resolution representation learning for human pose
  estimation.
\newblock In \emph{Proceedings of the IEEE/CVF Conference on Computer Vision
  and Pattern Recognition}, pp.\  5693--5703, 2019.

\bibitem[Vaswani et~al.(2017)Vaswani, Shazeer, Parmar, Uszkoreit, Jones, Gomez,
  Kaiser, and Polosukhin]{vaswani2017attention}
Ashish Vaswani, Noam Shazeer, Niki Parmar, Jakob Uszkoreit, Llion Jones,
  Aidan~N Gomez, Lukasz Kaiser, and Illia Polosukhin.
\newblock Attention is all you need.
\newblock In \emph{Advances in neural information processing systems}, pp.\
  5998--6008, 2017.

\bibitem[Wang et~al.(2011)Wang, Babenko, and Belongie]{wang2011end}
Kai Wang, Boris Babenko, and Serge Belongie.
\newblock End-to-end scene text recognition.
\newblock In \emph{2011 International Conference on Computer Vision}, pp.\
  1457--1464. IEEE, 2011.

\bibitem[Wang et~al.(2018)Wang, Liu, Liu, He, Lyu, Wu, Li, and
  Wang]{wang2018multi}
Yizhong Wang, Kai Liu, Jing Liu, Wei He, Yajuan Lyu, Hua Wu, Sujian Li, and
  Haifeng Wang.
\newblock Multi-passage machine reading comprehension with cross-passage answer
  verification.
\newblock \emph{arXiv preprint arXiv:1805.02220}, 2018.

\bibitem[Xiao et~al.(2018)Xiao, Wu, and Wei]{xiao2018simple}
Bin Xiao, Haiping Wu, and Yichen Wei.
\newblock Simple baselines for human pose estimation and tracking.
\newblock In \emph{Proceedings of the European conference on computer vision
  (ECCV)}, pp.\  466--481, 2018.

\bibitem[Yang \& Ramanan(2012)Yang and Ramanan]{yang2012articulated}
Yi~Yang and Deva Ramanan.
\newblock Articulated human detection with flexible mixtures of parts.
\newblock \emph{IEEE transactions on pattern analysis and machine
  intelligence}, 35\penalty0 (12):\penalty0 2878--2890, 2012.

\bibitem[You et~al.(2017)You, Xu, Xu, and Tao]{you2017learning}
Shan You, Chang Xu, Chao Xu, and Dacheng Tao.
\newblock Learning from multiple teacher networks.
\newblock In \emph{Proceedings of the 23rd ACM SIGKDD International Conference
  on Knowledge Discovery and Data Mining}, pp.\  1285--1294, 2017.

\bibitem[You et~al.(2020)You, Huang, Yang, Wang, Qian, and
  Zhang]{you2020greedynas}
Shan You, Tao Huang, Mingmin Yang, Fei Wang, Chen Qian, and Changshui Zhang.
\newblock Greedynas: Towards fast one-shot nas with greedy supernet.
\newblock In \emph{Proceedings of the IEEE/CVF Conference on Computer Vision
  and Pattern Recognition}, pp.\  1999--2008, 2020.

\bibitem[Zhang et~al.(2020)Zhang, Zhu, Dai, Ye, and Zhu]{zhang2020distribution}
Feng Zhang, Xiatian Zhu, Hanbin Dai, Mao Ye, and Ce~Zhu.
\newblock Distribution-aware coordinate representation for human pose
  estimation.
\newblock In \emph{Proceedings of the IEEE/CVF conference on computer vision
  and pattern recognition}, pp.\  7093--7102, 2020.

\end{thebibliography}
